\documentclass[sigconf]{acmart}
\usepackage{cleveref}
\crefname{figure}{Fig.}{Fig.}
\usepackage{graphicx}
\usepackage{amsmath}
\usepackage{amsfonts}
\usepackage{booktabs}
\usepackage{subcaption}
\usepackage{enumitem}
\usepackage{float}
\usepackage{multirow}
\usepackage{tikz}
\usepackage{booktabs}

\AtBeginDocument{%
  \providecommand\BibTeX{{%
    \normalfont B\kern-0.5em{\scshape i\kern-0.25em b}\kern-0.8em\TeX}}}

\setcopyright{acmlicensed}
\copyrightyear{2024}
\acmYear{2024}
\acmDOI{10.1145/3664647.3681337}

\begin{document}

\title{TimeNeRF: Building Generalizable Neural Radiance Fields across Time from Few-Shot Input Views}


\author{Hsiang-Hui Hung}
\orcid{0009-0004-7281-0763}
\affiliation{%
  \institution{National Yang Ming Chiao Tung University}
  \city{Hsinchu}
  \country{Taiwan}
  }
\email{sofina604.cs10@nycu.edu.tw}
\authornote{Both authors contributed equally to the paper}

\author{Huu-Phu Do}
\orcid{0009-0006-7327-9016}
\affiliation{%
  \institution{National Yang Ming Chiao Tung University}
  \city{Hsinchu}
  \country{Taiwan}
  }
\email{dohuuphu25.ee11@nycu.edu.tw}
\authornotemark[1]

\author{Yung-Hui Li}
\orcid{0000-0002-0475-3689}
\affiliation{%
  \institution{Hon Hai Research Institute }
  \city{Hsinchu}
  \country{Taiwan}
  }
\email{yunghui.li@foxconn.com}

\author{Ching-Chun Huang}
\orcid{0000-0002-4382-5083}
\affiliation{%
  \institution{National Yang Ming Chiao Tung University}
  \city{Hsinchu}
  \country{Taiwan}
  }
\email{chingchun@nycu.edu.tw}
\authornote{Ching-Chun Huang is the corresponding author.}
\renewcommand{\shortauthors}{Hsiang-Hui Hung, Huu-Phu Do, Yung-Hui Li, Ching-Chun Huang}

\newcommand{\CCH}[1]{{\color{black}#1}\normalfont} 
\begin{abstract}
We present TimeNeRF, a generalizable neural rendering approach for rendering novel views at arbitrary viewpoints and at arbitrary times, even with few input views. For real-world applications, it is expensive to collect multiple views and inefficient to re-optimize for unseen scenes. Moreover, as the digital realm, particularly the metaverse, strives for increasingly immersive experiences, the ability to model 3D environments that naturally transition between day and night becomes paramount. While current techniques based on Neural Radiance Fields (NeRF) have shown remarkable proficiency in synthesizing novel views, the exploration of NeRF's potential for temporal 3D scene modeling remains limited, with no dedicated datasets available for this purpose. To this end, our approach harnesses the strengths of multi-view stereo, neural radiance fields, and disentanglement strategies across diverse datasets. This equips our model with the capability for generalizability in a few-shot setting, allows us to construct an implicit content radiance field for scene representation, and further enables the building of neural radiance fields at any arbitrary time. Finally, we synthesize novel views of that time via volume rendering. Experiments show that TimeNeRF can render novel views in a few-shot setting without per-scene optimization. Most notably, it excels in creating realistic novel views that transition smoothly across different times, adeptly capturing intricate natural scene changes from dawn to dusk. 
\end{abstract}





\maketitle

\begin{figure*}
  \centering
  \includegraphics[scale=0.2]{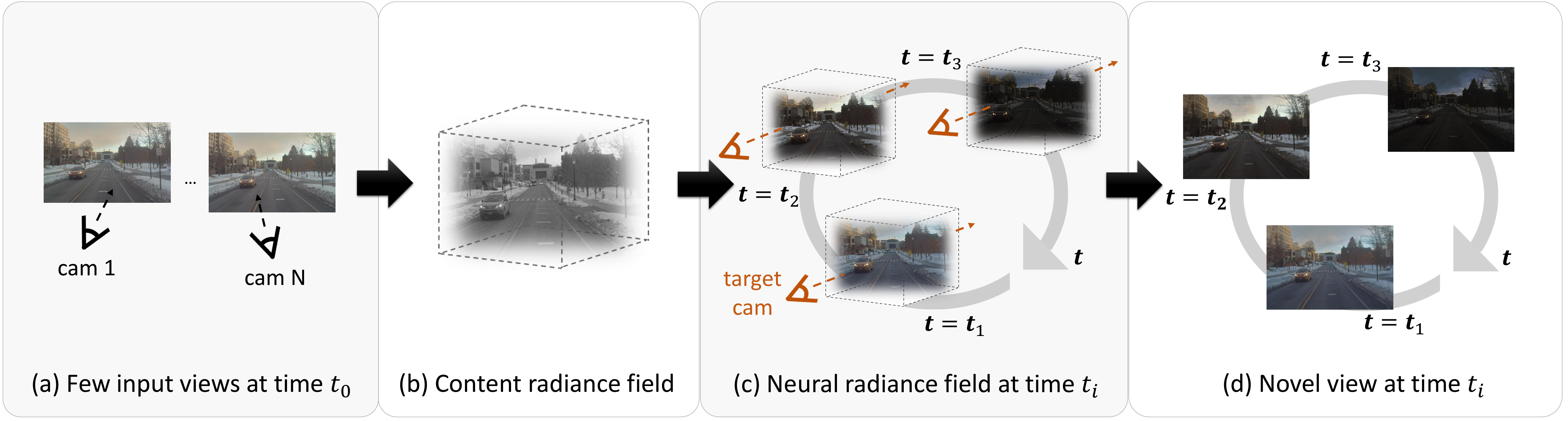}
  \caption{\textbf{Overview of TimeNeRF.} By inputting few input views (a), our method first constructs a content radiance field that filters out environmental changing factors (b), enabling us to obtain neural radiance fields at arbitrary times (c). Finally, our model seamlessly renders novel views with a smooth transition of time (d), providing an immersive and realistic experience.}
  \label{fig:buildTimeNeRF}
  \vspace{-5pt}
\end{figure*}

\section{Introduction}
\label{sec:intro}
\CCH{
Novel view synthesis (NVS), an essential challenge in computer vision, aims to synthesize unseen viewpoints from posed images. Its applications range from virtual reality (VR) and augmented reality (AR) to 3D scene reconstruction. Also, it’s a crucial technique to achieve the metaverse. 
The rise of neural rendering techniques, especially Neural Radiance Fields (NeRF) \cite{nerf} and its successors \cite{nerf++,instantngp,DVGO,NSVF}, has ushered in impressive progress in novel view synthesis. However, a notable drawback of these prior works is their reliance on per-scene optimization and the need for hundreds of different viewpoint images. In practical scenarios, input views are often limited, and re-optimizing the model for new scenes is inefficient. Moreover, to achieve complete immersion in virtual reality or the metaverse, creating environments that can transition seamlessly from day to night is essential. Nevertheless,
the exploration of NeRF’s potential for
temporal 3D scene modeling remains limited, with few dedicated
datasets available for this purpose.

In this paper, we present TimeNeRF, a framework designed to achieve novel view synthesis in three aspects: 1) a few-shot setting, 2) generalizing to new scenes captured under varying conditions, and 3) handling temporal transitions in 3D scenes.
Several studies are devoted to realizing novel view synthesis from sparse viewpoints \cite{Regnerf,infonerf,sinnerf,mixnerf} and also ensuring generalizability to previously unseen scenes \cite{pixelnerf,ibrnet,mvsnerf,neuray,GeoNeRF,visionnerf}. However, when it comes to rendering novel views at different times, the viable option for these methods is capturing an additional set of views at the desired time. On the other hand, NeRF-W \cite{nerfw} and Ha-NeRF \cite{hanerf} explore constructing neural radiance fields from images taken at various times and under different illuminations. Though capable of generating novel views at different times, they still depend on appearance codes from reference images and require many viewpoints and per-scene optimization. Recently, CoMoGAN \cite{comogan} introduces a continuous image translation model capable of altering an image's appearance to correspond to a different time. Yet, when integrated directly with a novel view synthesis model, it succumbs to the view inconsistency problem, a limitation highlighted in earlier works \cite{learnStyle,style3D,Huang22StylizedNeRF}.

Consequently, a unified 3D representation model that generalizes across different 3D scenes over time, especially in a few-shot setting, remains a challenging open question. To address this issue, we introduce TimeNeRF. Distinct from prior approaches, TimeNeRF synthesizes novel views from limited viewpoints without necessitating model retraining for previously unseen scenes. Moreover, by constructing novel time-dependent neural radiance fields, our method can render novel views at specific times without relying on reference images.

The key idea of TimeNeRF, as shown in \cref{fig:buildTimeNeRF}, is to first construct a content radiance field, then transform it into the neural radiance field of a specific moment by infusing the environmental change information relevant to the desired time into the content radiance field. Specifically, our approach involves a two-stage training process. The first stage focuses on disentangling content and environmental change factors, achieved through an image translation model. The second stage begins by extracting geometry information from a few input views via the appearance-agnostic geometry extractor. We build our model upon previous generalizable NeRF-based methods but introduce a significant adjustment: the cost volume is constructed from content features instead of being extracted from standard convolutional network features, allowing our model to handle various capture conditions. Next, the implicit scene network constructs a content radiance field based on the extracted geometry and features. We then predict the time and extract time-irrelevant features from the style feature obtained from a pre-trained style extractor. Finally, we create time-dependent radiance fields by incorporating the time and the extracted features into the content radiance field for novel view synthesis over time. 

To train the model, it is ideal to have a dataset with various capture views and different capture times. However, collecting dense inputs of a scene is expensive, especially considering our specific problem setup, which requires generating novel views at particular times. This means that, in addition to the substantial effort required to gather multiple views of the scene, we also need to collect data at different times throughout the day, which in practice becomes very costly. To overcome this, we train our model on both the Ithaca365 dataset \cite{ithaca}, which offers few different capture views but has limited time information, and the Waymo dataset \cite{waymo}, covering a range of capturing times (i.e., day, dusk, dawn, night). Since these datasets do not contain time labels, we train the model without relying on exact time data. Specifically, we leverage our two-stage training approach with the design of our network to map reference images into the time domain. The design also benefits the testing phase, allowing us to render novel views at any time by specifying the time code directly. Additionally, we introduce several loss functions to ensure that our model exhibits cyclic changes and that the transition results are adapted based on the inputs. We will explain the details in the ``Proposed Method'' section. In general, our main contributions are:
\begin{itemize}[itemsep=0pt, parsep=0pt, left=-0.05em]
    \item An extended novel setting for view synthesis over time, which is more practical for real-world applications.
    \item A novel time-dependent and NeRF-based approach that renders novel views at arbitrary times from few inputs and achieves generalizability to new scenes captured under any conditions.
    \item Extensive experiments showing TimeNeRF's capability to transition smoothly across different times of the day.
\end{itemize}
}
\vspace{-7pt}
\section{Related works}
\label{sec:related_work}

\CCH{
\subsection{Neural Radiance Field}

Neural Radiance Field (NeRF) \cite{nerf} revolutionized novel view synthesis by modeling 3D scenes as an implicit neural representation. It employs a multi-layer perceptron (MLP) to map 3D positions and camera directions to colors and densities. 
Subsequent works improve NeRF by combining implicit scene representation with traditional grid representation \cite{DVGO,evgnerf,NSVF,SNeRG,DIVeR,instantngp} or reducing samples taken along each ray \cite{donerf, adanerf2022,TermiNeRF,autoint2021}.
Nevertheless, these techniques are limited in generating views for different time points and face challenges with varied illumination.

NeRF-W \cite{nerfw} extends the NeRF model by introducing a learned appearance code for each image, allowing it to synthesize novel views at different times by inputting different appearance codes. The subsequent works \cite{hanerf,Block-NeRF,NeROIC,kplanes_2023} adopt a similar idea. However, these methods require per-scene optimization and a large number of training views. Moreover, they lack the ability to query the view at a specific time.

On the other hand, NeRF to dynamic scene \cite{d-nerf,DyNeRF,kplanes_2023,hypernerf,NeuralVolumes} aims to model a dynamic scene by learning a 4D implicit scene representation. The inputs of the MLP include not only 3D coordinates and direction but also time. While these approaches target short-range temporal changes, like moving people or objects, we emphasize modeling long-range time shifts, enabling realistic view rendering from day to night. Furthermore, these methods also require many image views and re-optimizing the model for each new scene.
\vspace{-5pt}
\subsection{Few-Shot NeRF}

Recently, three major approaches have been proposed to enable the synthesis of novel views from limited inputs. First, some methods \cite{dietNeRF,infonerf,Regnerf,sinnerf,freenerf,mixnerf} incorporate semantic and geometry regularizations to constrain the output color and density.  
Second, another line of works attempts to leverage additional depth information such as sparse 3D points \cite{dsnerf,depthpriorsnerf} or depth prediction from a pre-trained model \cite{sparsenerf}. 
Third, some other approaches \cite{pixelnerf,mvsnerf,ibrnet,neuray,GeoNeRF,rgbdNet,ContraNeRF,lirf,visionnerf} attempt to condition the model with features extracted from inputs, allowing it to be generalizable to new scenes. For example, MVSNeRF \cite{mvsnerf} proposes incorporating a 3D cost volume constructed from the extracted feature of input views \cite{mvsnet,cascade} with the NeRF model for assistance. This allows the model to develop geometry awareness of scenes and increases model generalization ability. Later works \cite{neuray,GeoNeRF,rgbdNet,ContraNeRF} also adopt this idea. GeoNeRF \cite{GeoNeRF} and NeuRay \cite{neuray} further take the occlusion problem into account by predicting the visibility of each source view. ContraNeRF \cite{ContraNeRF} introduces a geometry-aware contrastive loss to improve generalization in the synthetic-to-real setting. However, the methods above do not support rendering a novel view at arbitrary times.

\subsection{Natural Image Synthesis with Time Translation}

Transferring an image to another time zone, for example, from day to night or summer to winter, usually involves style transfer methods. The typical approach is to disentangle content and style information, where style corresponds to the time-specific characteristics. Subsequently, an image is transferred by replacing its original style with the desired time-related style, which is extracted from a reference image \cite{DRIT,DRIT_plus,tsit,UnifyingG}. However, this kind of method is limited to domain transfer, lacks the ability to query a specific time, and cannot achieve continuous translation across different time periods. Earlier works \cite{DNI,DLOW,smit} on the continuous image translation task assume linear domain manifolds, which may not be suitable for daytime translation. Daytime translation should be cyclic, allowing for translation from day to dusk, dusk to night, night to dawn, and so forth. CoMoGAN \cite{comogan} proposes the first continuous image translation framework, which enables cyclic or non-linear translation through the design of the functional instance layer and the guidance of a non-neural physical model. Besides, some research efforts are focusing on timelapse generation \cite{End-To-End_Time-Lapse,HiDT,Time-of-Day,Time_Flies,Landscape_Animation}, aiming to generate a timelapse video based on a single source image. However, none of these works are equipped for time transitions in a 3D space, which enhances the overall immersive experience in digital realms, especially in the realm of the metaverse.
}

\CCH{
\section{Proposed Method}
\label{sec:proposed_method}

\begin{figure*}
  \centering
  \includegraphics[scale=0.268]{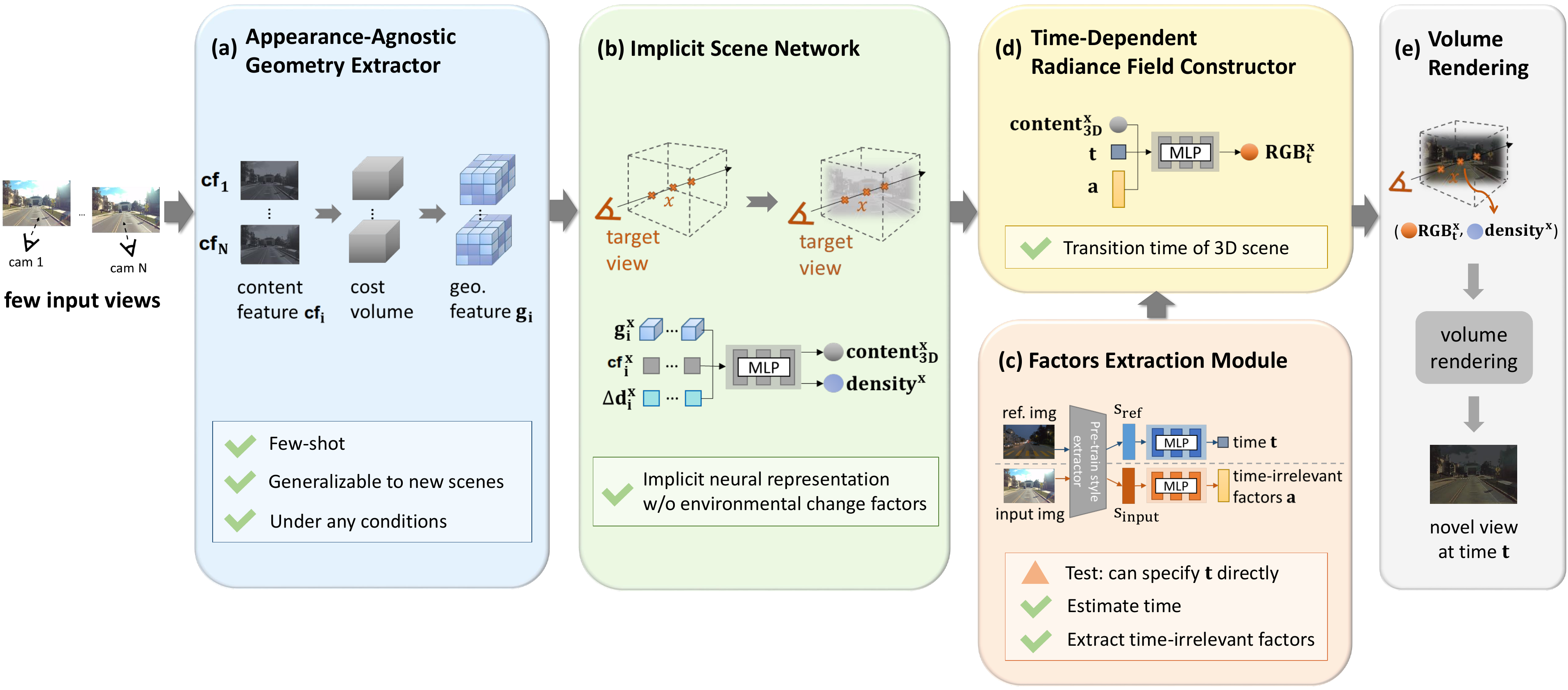}
  \caption{\textbf{Architecture Overview.} The proposed framework comprises five main parts: (a) The appearance-agnostic geometry extractor is designed to extract geometry features for each input view. The design of this module allows the model to operate in a few-shot setting without the need for per-scene optimization. It constructs cost volumes and works under various capture conditions by utilizing the content features (\cref{subsec:GeometryExtractor}). (b) For each sample point \(\mathbf{x}\), the implicit scene network predicts its corresponding content feature and density by aggregating geometry features, content features, and viewing directions, thus constructing a content radiance field (\cref{subsec:ImplicitSceneNetwork}). (c) The factor extraction module predicts both the time and time-irrelevant features. Note that time prediction is only required during training (\cref{subsec:FactorsExtractionModule}). (d) The time-dependent radiance field constructor transforms the content radiance field into the time-dependent radiance field based on the information from the factor extraction module (\cref{subsec:Time-DependentRadianceFieldConstructor}). (e) Finally, a novel view at time \(t\) is rendered via standard volume rendering.}
  \label{fig:framework}
\vspace{-5pt}
\end{figure*}

Our goal is to synthesize novel views at any given time by learning implicit scene representations across time from sparse input views that are not captured at the desired moment. We begin by reviewing NeRF and highlighting the distinctions in our approach (\cref{subsec:Preliminaries}). After detailing our training process (\cref{subsec:TrainingProcess}), we introduce our proposed framework. As depicted in \cref{fig:framework}, our framework comprises five main components: 1) extracting geometry features for each source view (\cref{subsec:GeometryExtractor}), 2) constructing a content radiance field (\cref{subsec:ImplicitSceneNetwork}) which stores densities and content features, whose environmental change factors have been excluded, 3) estimating time and extracting time-irrelevant factors (\cref{subsec:FactorsExtractionModule}), 4) transferring content features at 3D locations to RGB colors of the specific time point and complete a time-dependent radiance field (\cref{subsec:Time-DependentRadianceFieldConstructor}), and 5) rendering a novel view at the desired time through volume rendering (\cref{subsec:Preliminaries}). Finally, the proposed loss functions are designed to further improve cyclic changes and enhance the content adaptation based on the input data (\cref{subsec:LossFunctions}). Due to space constraints, an in-depth description of the network architecture and extra discussion are relegated to the supplementary material. The codebase is ready to open and will be made publicly available in due course.   

\subsection{Preliminaries}
\label{subsec:Preliminaries}

First of all, we briefly review the idea of NeRF \cite{nerf}. NeRF optimizes an MLP, whose input consists of 3D spatial location \(\mathbf{x}\) and viewing direction \(\mathbf{d}\), and whose output corresponds to colors \(\mathbf{c}\) and densities \(\sigma\), to represent a scene implicitly. In other words, it aims to learn a continuous function: \((\mathbf{c}, \sigma)  = \ F_\theta(\mathbf{x}, \mathbf{d})\). To render a pixel in an image, NeRF first samples \(M\) points on the corresponding 3D ray \(\mathbf{r}\) and obtains the colors and densities of these sample points. Then, NeRF renders the 2D-pixel color using the volume rendering technique; the formula is described as follows.

\begin{equation} \label{eq:volumerendering}
\vspace{-5pt}
    \hat{C}(\mathbf{r})=\sum_{i=1}^M T_i\left(1-\exp \left(-\sigma_i \delta_i\right)\right) \mathbf{c}_i. 
\end{equation}
\begin{equation} \label{eq:Ti}
    T_i=\exp \left(-\sum_{j=1}^{i-1} \sigma_j \delta_j\right),
\end{equation}
where \(M\) is the number of sample points along a ray \(\mathbf{r}\) and \((\mathbf{c}_i,\sigma_i)\) represent the color and density of point \(\mathbf{x}_i\), respectively. \(\delta_i\) denotes the distance of the adjacent sample points.

In our method, we introduce novel modifications to the original NeRF. First, following IBRNet \cite{ibrnet}, \(\delta_i\) in \cref{eq:Ti} is removed for better generalizability.
Second, to enable generalizability, we replace the position \(\mathbf{x}\) in the model's input with the feature \(\mathbf{f}^{\mathbf{x}}\), which is derived from the interpolation of features extracted from input views at the projection pixels corresponding to the 3D position \(\mathbf{x}\).
Third, we aim to render a novel view at the arbitrary time \(t\). Therefore, our TimeNeRF is designed to learn a continuous function: \( (\mathbf{c}_t, \sigma)=\ F_\phi(\mathbf{f}^{\mathbf{x}}, \mathbf{d}, t) \), where \(\mathbf{c}_t\) is the color in direction \(\mathbf{d}\) at position \(\mathbf{x}\) and time \(t\) within a scene. 

\subsection{Training Process}
\label{subsec:TrainingProcess}

We employ a two-stage training process to effectively disentangle content and environmental change factors and subsequently construct implicit scene representations. In Stage 1, we achieve feature disentanglement by leveraging existing style translation models comprising three modules: content extractor, style extractor, and generator. Specifically, we train DRIT++ \cite{DRIT_plus} on the Ithaca365 dataset \cite{ithaca}, which contains images across varied weather and nighttime conditions. To enhance the content extractor in DRIT++, instead of only using the content extractor's output, we extract/select content features from 3 different convolutional layers corresponding to different semantic levels for styled image generation. This strategy refines early-layer content extraction, as these features are utilized to generate images, thereby boosting the model's overall content extraction ability.

After Stage 1 training, DRIT++ offers disentangled content features and enables the creation of stylized images as pseudo ground truth for Stage 2 training. In Stage 2, we train our TimeNeRF (detailed in \cref{subsec:GeometryExtractor} to \cref{subsec:Time-DependentRadianceFieldConstructor}) using input (source) views from the Ithaca365 dataset and using reference images from the Waymo dataset \cite{waymo}. By merging source images' content features with reference images' style features for training guidance, we can learn the implicit 3D scene representations across time \(F_\phi(\cdot)\). The reason behind using the two datasets is elaborated in \cref{subsec:Dataset}.

\subsection{Appearance-Agnostic Geometry Extractor}
\label{subsec:GeometryExtractor}

Given input views \(\{I_i\}_{i=1}^{N}\) and their corresponding camera parameters \(\{E_i, K_i\}_{i=1}^{N}\), we extract content features at three levels \(\{cf_i^{(l)}\}_{i=1}^{N}\) (i.e., \(l=0,1,2\) and \(l=0\) is the lowest level feature) for the \(N\) input views using the pre-trained content extractor. The content features exclude environmental change factors, such as temporal and weather factors, because the image style has been removed using the pre-trained model. This allows us to handle input views under varying capture conditions. Next, for each input view \(v\), we select \(S\) nearby views and construct cost volumes \cite{mvsnet,GeoNeRF,neuray,cascade} at three levels \(\{C_v^{(l)}\}_{l=0}^{2}\) by warping the content features of the \(S\) nearby views to align with view \(v\) using the corresponding camera parameters. Finally, the cost volumes \(\{C_v^{(l)}\}_{l=0}^{2}\) are put into a 3D-Unet to obtain geometry features \(\{g_v^{(l)}\}_{l=0}^{2}\) for the input view \(v\). In the following, we denote \(\{cf_i^{(l)}\}_{l=0}^2\) as \(cf_i\) and \(\{g_i^{(l)}\}_{l=0}^2\) as \(g_i\).

\subsection{Implicit Scene Network}
\label{subsec:ImplicitSceneNetwork}
After obtaining geometry features, we may use them to predict colors and densities (\cref{subsec:Preliminaries}). However, this would lose the ability of modeling style change over time. Instead, we propose constructing a content radiance field, an implicit scene representation without environmental change factors. To achieve this, inspired by GeoNeRF \cite{GeoNeRF}, we aggregate the extracted geometry features to predict each sample point's density and content features. For a sample point  \(\mathbf{x} \in \mathbb{R}^3\) along a ray \(\mathbf{r}\), we interpolate geometry features \(\{g_i\}_{i=1}^{N}\) and 2D content features \(\{cf_i\}_{i=1}^{N}\), as opposed to the 2D features used in GeoNeRF, to get its corresponding features \(\{g_i^\mathbf{x}\}_{i=1}^{N}\) and \(\{cf_i^\mathbf{x}\}_{i=1}^{N}\) at point \(\mathbf{x}\). These are then aggregated to estimate the density and content feature for \(\mathbf{x}\). The procedure is described below.

First, we aggregate features via fully-connected layers and multi-head attention layers \cite{transformer}, which facilitate the exchange of information between different views.
\begin{equation} \label{eq:H()}
    \sigma^\mathbf{x}, \{\tilde{w}_i^\mathbf{x}\}_{i=1}^N = H(\{cf_i^\mathbf{x}\}_{i=1}^N, \{g_i^\mathbf{x}\}_{i=1}^N),
\end{equation}
where \(H\) consists of fully-connected layers and multi-head attention layers. \(\sigma^{\mathbf{x}}\) is the density of \(\mathbf{x}\) and \(\{\tilde{w}_i^\mathbf{x}\}_{i=1}^N\) denotes the enhanced features of \(\mathbf{x}\) for each input view. To obtain the 3D content feature \(\xi^\mathbf{x}\) for a point \(\mathbf{x} \in \mathbb{R}^3\), we predict the weights of input views \(\{w_i^{\mathbf{x},v}\}_{i=1}^N\) and use them to calculate the weighted sum of 2D content features \(\{cf_i^\mathbf{x}\}_{i=1}^{N}\). That is 
\begin{equation} \label{eq:weight}
    \{w_i^{\mathbf{x},v}\}_{i=1}^{N} = \text{softmax}\left(\text{MLP}_w\left(\{\tilde{w}^\mathbf{x}_i \,||\, {\Delta d}_i^\mathbf{x}\}_{i=1}^{N}\right)\right),
\end{equation}
\begin{equation}
\label{eq:content3d}
    \xi^\mathbf{x} = \sum_{i=1}^{N} w_i^{\mathbf{x},v} \cdot cf_i^\mathbf{x},
\end{equation}
where \({\Delta d}_i^\mathbf{x}\) represents the direction difference between the query view \(v\) and input view \(i\) by computing the cosine similarity, and the concatenation is denoted by \(||\).
We predict the weights via an MLP that considers both the direction difference \({\Delta d}_i^\mathbf{x}\) and \(\tilde{w}_i^\mathbf{x}\) in \cref{eq:weight}.
\(\xi\) stands for \(\{{\xi}^{(l)}\}_{l=0}^2\). Each individual 3D content feature \(\xi^{(l)}\) is computed from the 2D content features \(\{cf_i^{(l)}\}_{i=1}^{N}\) at level \(l\) by \cref{eq:content3d}.

\subsection{Factors Extraction Module}
\label{subsec:FactorsExtractionModule}
To infuse environmental change information into the content radiance field, one intuitive approach is to use the style feature extracted from the pre-trained style extractor. The style feature inherently contains both time-relevant and time-irrelevant environmental factors since the translation model is trained to transfer images to diverse weather conditions and varying times. However, to purely manipulate the timing of radiance fields, it's imperative to separate time-related and time-irrelevant information further from the style feature. To this end, we employ two MLPs: \(g_t(\cdot)\) predict time information \(t\) and \(g_a(\cdot)\) extracts time-irrelevant features \(a\) from an image. During training, we extract \(t\) from a reference image and \(a\) from input images.

To encapsulate the cyclical progression of a day, we design \(g_t(\cdot)\) to map the style feature of a reference image onto the interval \([0, 2\pi)\), symbolizing the entire 24-hour cycle. The mapping is trained in an unsupervised manner, utilizing our pseudo stylized image loss \(L_{stylemse}\) (\cref{eq:stylemse}) to guide the model in extracting time-related information from reference images and mapping it to the time range \([0,2\pi)\). By this design, without time labels for training, we can still simulate the variations occurring at different times within a day. After training, the transition of a day is encoded within \([0,2\pi)\), enabling us to recreate scenes at a desired time \(t\) by tuning \(t\) during testing.



\subsection{Time-Dependent Radiance Field Constructor}
\label{subsec:Time-DependentRadianceFieldConstructor}

\begin{figure}
  \centering
  \includegraphics[scale=0.43]{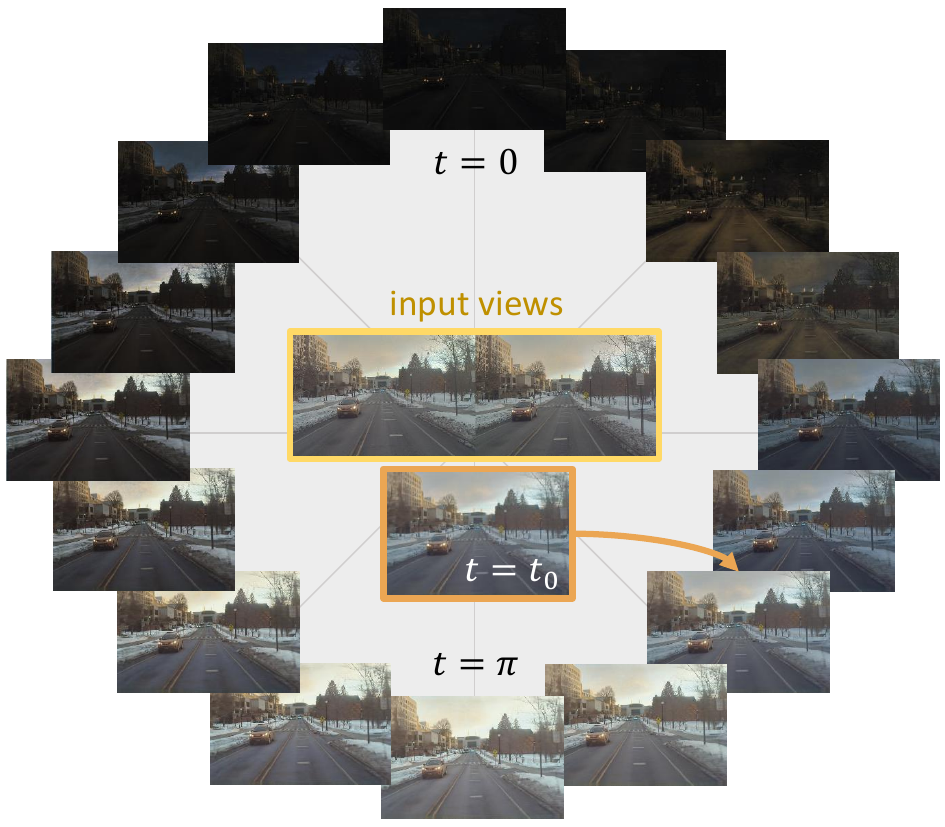}
  \caption{\textbf{Novel view synthesis across times.} The images in the yellow box represent the two input views of a test scene. The images around the circle are novel views at different times. The image in the orange box is synthesized for the time of input views \(t_0\) (\cref{eq:t0}), whose image style is consistent with input views.  }
  \label{fig:cycle}
\end{figure}

The objective of time-dependent radiance field constructor ``\( T \)'' is to transform the 3D content feature \(\xi^\mathbf{x}\) into a time-variant color \(\mathbf{c}_t^\mathbf{x}\) for a point \(\mathbf{x}\). This involves fusing \(\xi^\mathbf{x}\) with time-irrelevant environmental features \(a\) of input views and time \(t\). Essentially, we learn a mapping: \(\mathbf{c}_t^\mathbf{x}=T(\xi^\mathbf{x},a,t)\). 

The design of \(T\) is based on two main ideas: Firstly, we embed \(t\) into \((cos(t),sin(t))\), ensuring uniqueness for each value in \([0, 2\pi)\). This, in conjunction with \(L_{\Delta t}\) (\cref{eq:deltat}), facilitates cyclic changes. Secondly, to ensure the model focuses on learning variations based on time, we introduce a two-branch network. The first branch, fusing \(\xi\) with \(t\), acts as the template for the change over time. The second branch, merging \(\xi\) with both \(t\) and \(a\), is to further tune the color according to time-irrelevant features \(a\).

The network (detailed in Supplementary) is designed in a coarse-to-fine manner. We exploit content features at three different levels \(\{\xi^{(l)}\}_{l=0}^2\). The predicted color \(\mathbf{c}_t\) is refined progressively by integrating high-level features down to low-level information. Specifically, we obtain the feature of branch 1, \(f_1\), by combining content features from three levels with time \(t\). Meanwhile, we derive the feature of branch 2, \(f_2\), by integrating content features from three levels with both time \(t\) and time-irrelevant features \(a\). Finally, we combine \(f_1\) and \(f_2\) to produce the color \(\mathbf{c}_t\).

\subsection{Loss Functions}
\label{subsec:LossFunctions}

\noindent\textbf{MSE loss.} The mean square error loss \(\mathcal{L}_{mse}\) in \cref{eq:mse} ensures accurate 3D scene construction in our model. Different from the colors \(\mathbf{c}_t\) described in \cref{subsec:Time-DependentRadianceFieldConstructor}, the predicted colors \(\mathbf{c}\) in \cref{eq:c} for calculating \(\hat{C}(\mathbf{r})\) in this loss are derived from the weighted sum of the original input image colors \(\{I_i\}_{i=1}^N\), aiding in the learning of densities \(\sigma\) (\cref{eq:H()}) and weights \(\{w_i\}_{i=1}^N\) (\cref{eq:weight}).
\begin{equation}  
\label{eq:c}
    \mathbf{c^\mathbf{x}}=\sum_{i=1}^{N} w_i^{\mathbf{x},v} \cdot I_i^\mathbf{x}.
\end{equation}
\begin{equation}
\label{eq:mse}
    \mathcal{L}_{mse}=\frac{1}{|R|}\sum_{\mathbf{r} \in R}\|\hat{C}(\mathbf{r})-C(\mathbf{r})\|_2^2,
\end{equation}
where \(R\) is the set of rays in each training batch. \(I_i^{\mathbf{x}}\) is the projection color of point \(\mathbf{x}\) to image \(I_i\).  \(\hat{C}(\mathbf{r})\) is the predicted color computed by \cref{eq:volumerendering} using color \(\mathbf{c}\) in \cref{eq:c}; \({C}(\mathbf{r})\) is the ground truth color of the target view.
\\[3.5pt]
\noindent\textbf{Pseudo stylized loss.} The pseudo stylized loss \(\mathcal{L}_{stylemse}\) aims to guide the model to learn temporal transition by minimizing the difference between our predicted time-dependent pixel color and the pseudo ground truth colors.
\begin{equation}
\label{eq:stylemse}
    \mathcal{L}_{stylemse}=\frac{1}{|R|}\sum_{\mathbf{r} \in R}\|\hat{C}_t(\mathbf{r})-C_{pseudo}(\mathbf{r})\|_2^2,
\end{equation}
where \(\hat{C}_t(\mathbf{r})\) is the predicted color computed by \cref{eq:volumerendering} using \(\mathbf{c}_t\) illustrated in \cref{subsec:Time-DependentRadianceFieldConstructor}, and \(C_{pseudo}(\mathbf{r})\) is the pixel color of the pseudo ground truth, which is the style-transferred output via the pre-trained translation model (DRIT++) given the target view and reference image.
\\[3.5pt] 
\noindent\textbf{Loss term of \(\Delta t\).} Besides the design where we transform time \( t \) into \( ( \cos(t), \sin(t) ) \) to ensure cyclic changes, we introduce the \(\Delta t\) loss. This loss function in \cref{eq:deltat} leverages a small MLP, denoted as \( D \), to estimate the time difference based on color variations. The predicted time difference is then compared to the pseudo time difference \(\Delta t\):
\begin{equation}
    \Delta t = 
        \begin{cases} 
        |t - t'| & \text{if } |t - t'| \leq \pi \\
        2\pi - |t - t'| & \text{if } |t - t'| > \pi.
        \end{cases}
        \quad
\end{equation}
\begin{equation}
\label{eq:deltat}
    \mathcal{L}_{\Delta t}=\|D(\mathbf{c}_t,\mathbf{c}_{t'})-\Delta t\|_2^2,
\end{equation}
where \(t\) and \(t'\) are randomly sampled from \([0,2\pi)\), and \(\mathbf{c}_t\) and \(\mathbf{c}_{t'}\) are the predicted colors at times \(t\) and \(t'\), respectively. By minimizing this loss, our color predictions are guaranteed to align with time-based changes.
\\[3.5pt]
\noindent\textbf{Reconstruction loss. }This loss is to ensure that rendered views match the original appearance when the input time aligns with the capture time. This further allows the rendered views of different times to adapt to the inputs. Particularly,
\begin{equation}
\label{eq:t0}
    t_0=g_t(s_{I}), ~ and
\end{equation}
\begin{equation}
\label{eq:Lt0}
    \mathcal{L}_{t_0rec}=\frac{1}{|R|}\sum_{\mathbf{r} \in R}\|\hat{C}_{t_0}(\mathbf{r})-C(\mathbf{r})\|_2^2,
\end{equation}
where \(t_0\) is derived from predicting the time of an input view \(I\) (\cref{eq:t0}) and \(s_I\) is the style feature of \(I\).
\\[3.5pt]
\noindent\textbf{Total loss.} The total loss is as follows:
\begin{equation}
    \mathcal{L}_{total}=\mathcal{L}_{mse}+\lambda_1\mathcal{L}_{stylemse}+\lambda_2\mathcal{L}_{\Delta t}+\lambda_3\mathcal{L}_{t_0rec},
\end{equation}
where we set \(\lambda_1=0.5\), \(\lambda_2=0.01\), and \(\lambda_3=0.5\).

}

\section{Experiments}
\label{sec:experiment}

\subsection{Datasets}
\label{subsec:Dataset}

\begin{figure*}
  \centering

  \begin{subfigure}{\linewidth}
    \begin{tabular}{@{}cc@{}}
      \addlinespace[-2.5mm]
      \parbox{12.5mm}{~} & \raisebox{0.\height}{\includegraphics[scale=0.433]{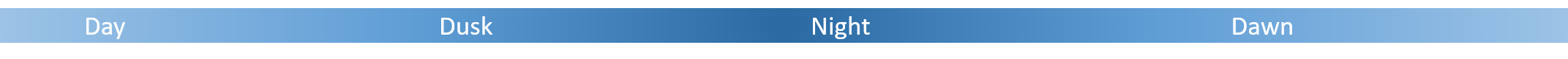}}\\
      \addlinespace[-2.8mm]
    \end{tabular}
  \end{subfigure}
  
  \begin{subfigure}{\linewidth}
    \begin{tabular}{@{}cc@{}}
      \raisebox{-0.5\height}{\parbox{12.5mm}{\centering\small ~~DRIT++* \cite{DRIT_plus}}} & \raisebox{-0.5\height}{\includegraphics[scale=0.433]{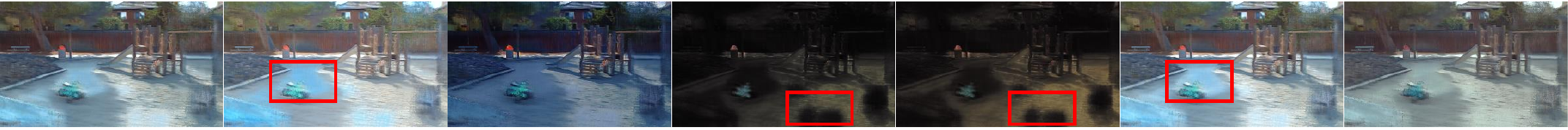}}
    \end{tabular}
  \end{subfigure}

  \begin{subfigure}{\linewidth}
    \begin{tabular}{@{}cc@{}}
      \raisebox{-0.5\height}{\parbox{12.5mm}{\centering\small HiDT* \cite{HiDT}}} & \raisebox{-0.5\height}{\includegraphics[scale=0.433]{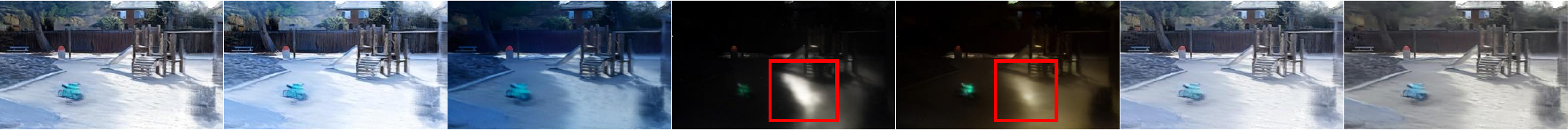}}
    \end{tabular}
  \end{subfigure}
  
  \begin{subfigure}{\linewidth}
    \begin{tabular}{@{}cc@{}}
      \raisebox{-0.5\height}{\parbox{12.5mm}{\centering\small CoMoGAN* \cite{comogan}}} & \raisebox{-0.5\height}{\includegraphics[scale=0.433]{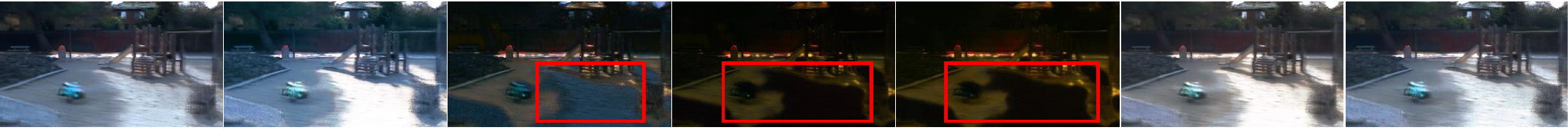}}
    \end{tabular}
  \end{subfigure}

  \begin{subfigure}{\linewidth}
    \begin{tabular}{@{}cc@{}}
      \parbox{12.5mm}{\centering\small Ours} & \raisebox{-0.5\height}{\includegraphics[scale=0.433]{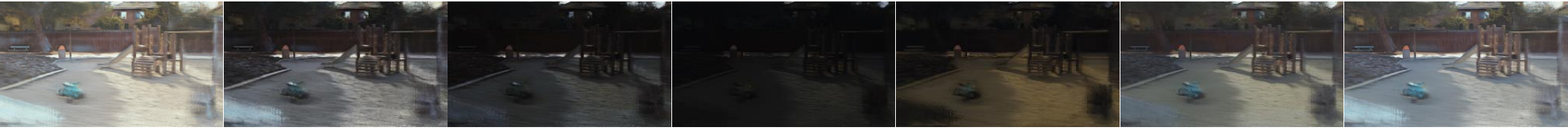}}
    \end{tabular}
  \end{subfigure}
  \caption{\textbf{Comparison of view synthesis across time.} We generate novel views at 7 different times to show the cyclic changes of a day from 3 input views. Unlike our approach, other methods first need to utilize the view synthesis model to render the novel view before executing time transitions, denoted by *. Besides, these methods may produce color bias and incorrect dark areas on the ground.}
  \label{fig:nvs+t_compare}
\end{figure*}

\begin{figure*}
  \centering
  
  \begin{tabular}{@{}c@{\hspace{0.4em}}c@{\hspace{0.4em}}c@{\hspace{0.4em}}c@{\hspace{0.4em}}c@{}}
    \small Ground truth & 
    \small (a) w/o \(L_{t_0rec}\) & 
    \small (b) w/o \(L_{\Delta t}\) & 
    \small (c) One-branch & 
    \small (d) Ours \\
    \raisebox{-0.5\height}{\includegraphics[scale=0.25]{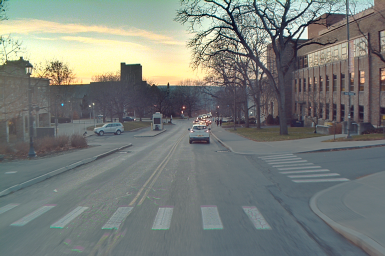}} &
    \raisebox{-0.5\height}{\includegraphics[scale=0.37]{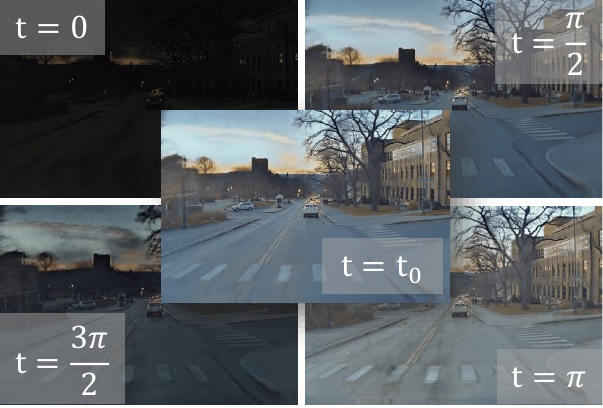}} & 
    \raisebox{-0.5\height}{\includegraphics[scale=0.37]{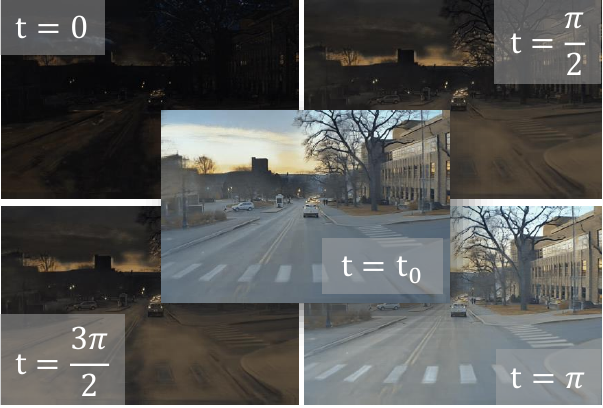}} &
    \raisebox{-0.5\height}{\includegraphics[scale=0.37]{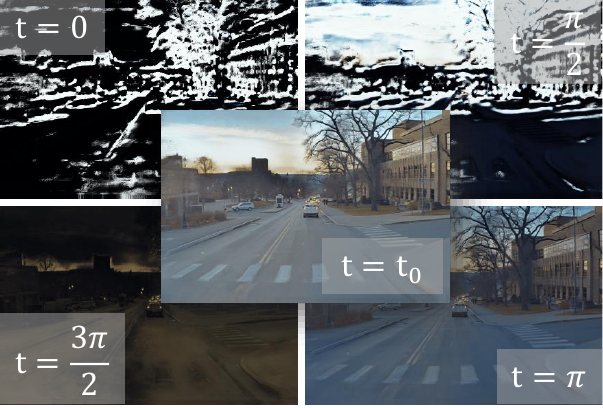}} &
    \raisebox{-0.5\height}{\includegraphics[scale=0.37]{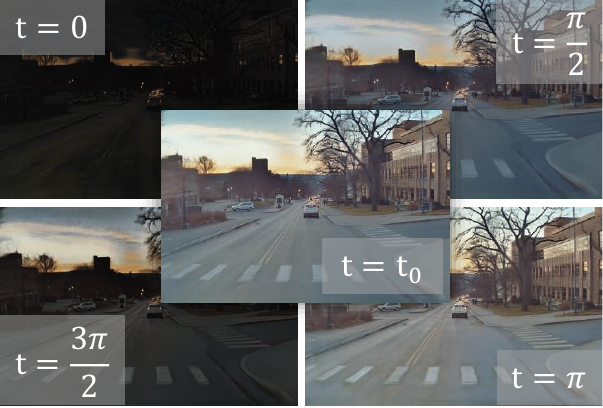}}
  \end{tabular}
  
  \caption{\textbf{Ablation study.} (a) shows the result when \(L_{t_0rec}\) is omitted (\cref{eq:Lt0}) (b) illustrates the outcome without \(L_{\Delta t}\) (\cref{eq:deltat}). (c) depicts the result when the two-branch network is not utilized (\cref{subsec:Time-DependentRadianceFieldConstructor}). Finally, (d) is the result from our complete framework.}

  \label{fig:ablation}
    \vspace{-5pt}
\end{figure*}

Collecting a dataset with different views and diverse capture times is challenging and unavailable. To address this, we train our model using both the Ithaca365 dataset \cite{ithaca} and the Waymo dataset \cite{waymo}. Ithaca365 provides different capture views but limited time variations, while Waymo offers images from different times (day, dusk, dawn, and night) on sunny days.
From Ithaca365, which includes conditions like sunny, cloudy, rainy, snowy, and nighttime, we randomly selected 2180 scenes with 3 distinct views. This allows TimeNeRF to be versatile for various conditions during testing. On the other hand, reference images for learning time information are taken from Waymo. In all experiments, TimeNeRF is trained with the aforementioned dataset configurations. For testing, we evaluate the model's generalizability using scenes outside the training set, as well as the T\&T dataset \cite{t&t} and the LLFF dataset \cite{llff}.
\vspace{-10pt}
\subsection{Implementation Details}
\label{subsec:ImplementationDetails}
During training, we use \(N=2\) input views due to the Ithaca365 dataset's limitations of having only 4 distinct viewpoints per scene, yet 2 of them are nearly overlapping. However, our model is flexible enough to handle more views during testing. We train TimeNeRF over 15 epochs, sampling 1024 rays as the training batch. The number of sample points \(M\) is set to 128. The Adam optimizer \cite{adam} is applied with a \(5\times10^{-4}\) learning rate and a cosine scheduler without restarting the optimizer \cite{sgd}. In the testing phase, we neither re-train nor fine-tune our model for new scenes.



\vspace{-10pt}

\subsection{Novel View Synthesis Across Time}
\label{subsec:NovelViewSynthesisAcrossTime}

\begin{figure}
  \centering
  
  \begin{subfigure}{\linewidth}
    \begin{tabular}{@{}cc@{}}
       & Target view 1~~~~~~~~~~~~~~~~~\hspace{1.5cm}Target view 2\\
      \parbox{12.5mm}{\centering\small Ground \\ Truth} & \raisebox{-0.5\height}{\includegraphics[scale=0.25]{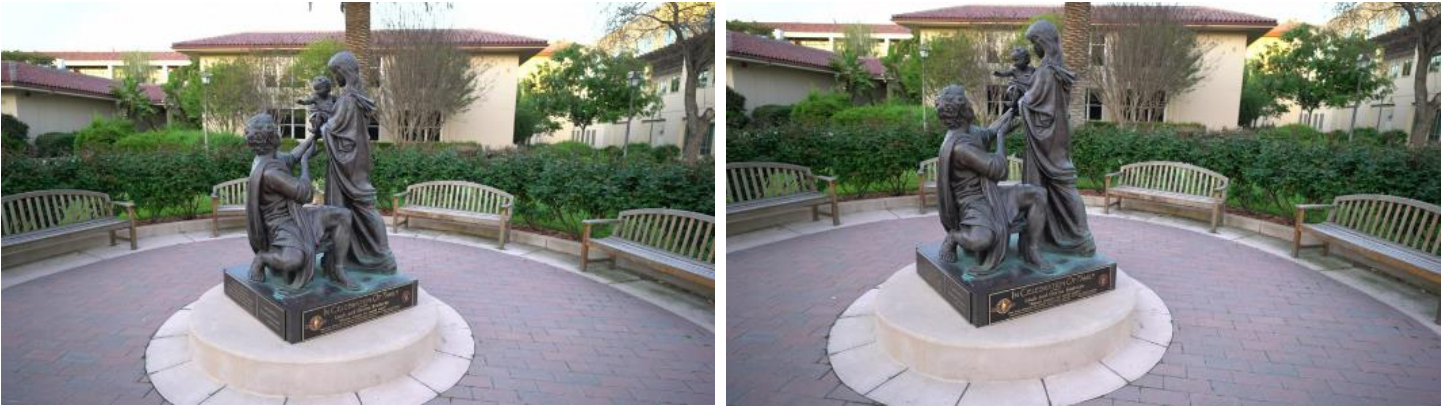}}
    \end{tabular}
  \end{subfigure}
  
  \begin{tikzpicture}
    \draw [dashed] (0,0) -- (\linewidth,0);
  \end{tikzpicture}
  
  \begin{subfigure}{\linewidth}
    \begin{tabular}{@{}cc@{}}
      \raisebox{-0.5\height}{\parbox{12.5mm}{\centering\small ~~DRIT++*\\\cite{DRIT_plus}}} & \raisebox{-0.5\height}{\includegraphics[scale=0.25]{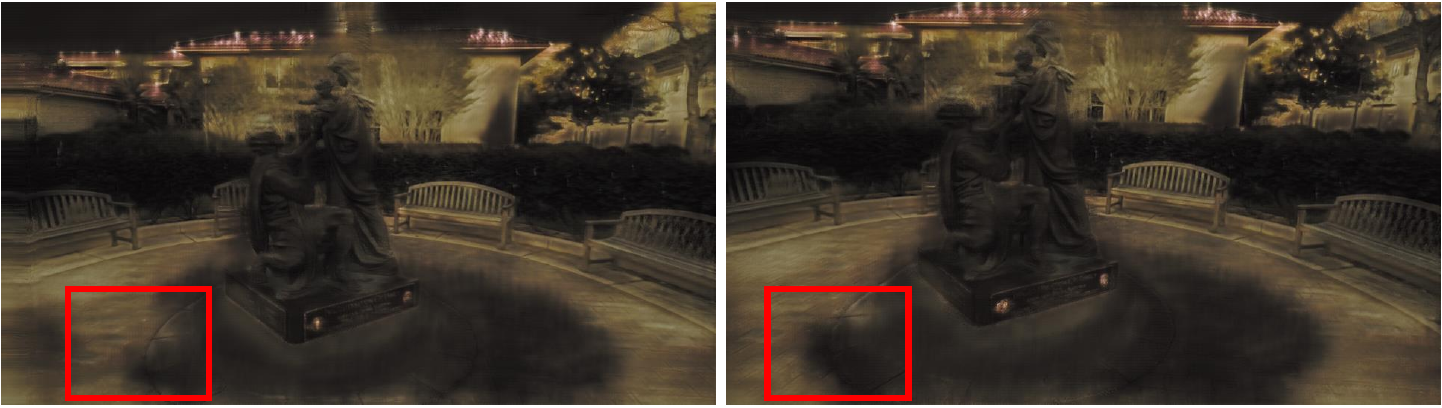}}
    \end{tabular}
  \end{subfigure}

  \begin{subfigure}{\linewidth}
    \begin{tabular}{@{}cc@{}}
      \raisebox{-0.5\height}{\parbox{12.5mm}{\centering\small HiDT*\\\cite{HiDT}}} & \raisebox{-0.5\height}{\includegraphics[scale=0.25]{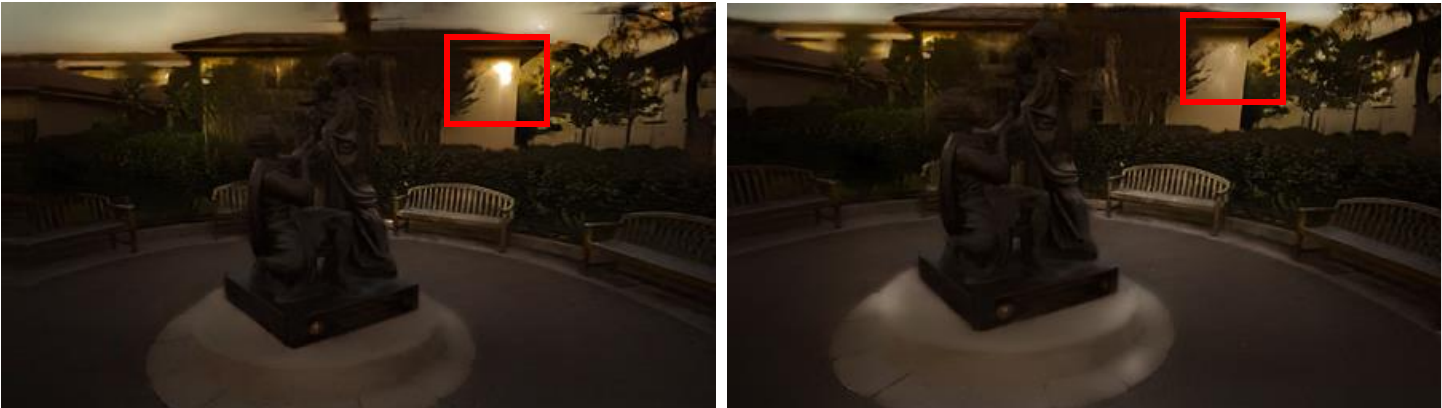}}
    \end{tabular}
  \end{subfigure}
  
  \begin{subfigure}{\linewidth}
    \begin{tabular}{@{}cc@{}}
      \raisebox{-0.5\height}{\parbox{12.5mm}{\centering\small CoMoGAN*\\\cite{comogan}}} & \raisebox{-0.5\height}{\includegraphics[scale=0.25]{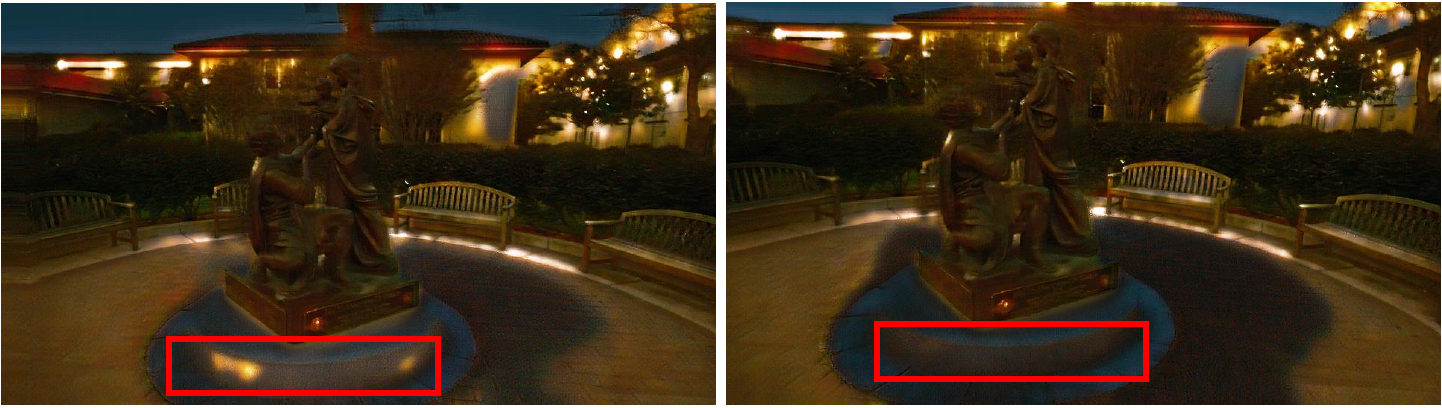}}
    \end{tabular}
  \end{subfigure}

  \begin{subfigure}{\linewidth}
    \begin{tabular}{@{}cc@{}}
      \parbox{12.5mm}{\centering\small Ours} & \raisebox{-0.5\height}{\includegraphics[scale=0.25]{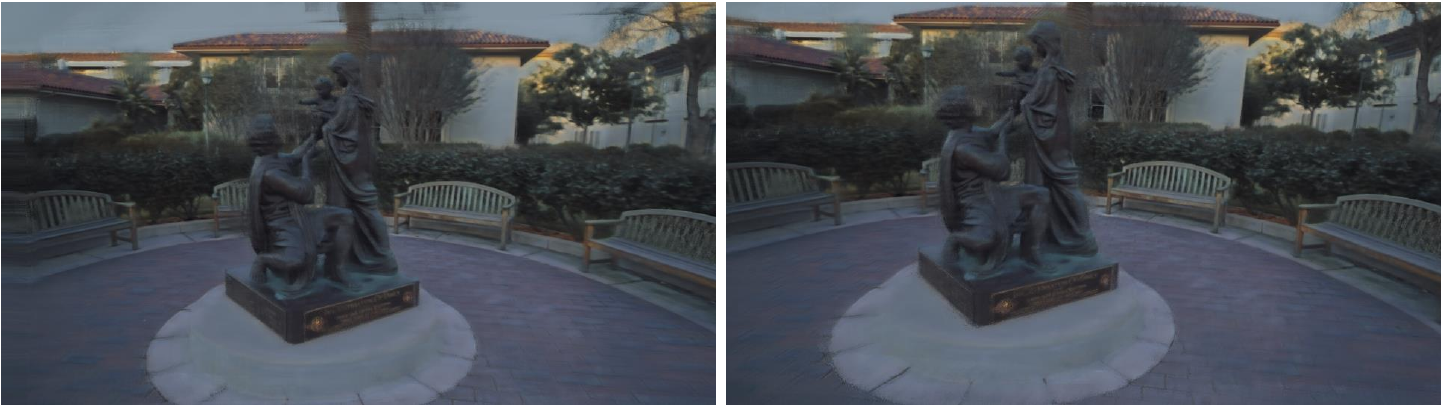}}
    \end{tabular}
  \end{subfigure}
  
  \caption{\textbf{View inconsistency issue.} We render 2 target views corresponding to the same time using the same set of 3 input views. The red boxes highlight the inconsistent regions between views. Among these results, our method produces more consistent views. }
  \label{fig:viewinconsist_compare}
  \vspace{-10pt}
\end{figure}

To show TimeNeRF's capability of rendering novel views over varying times, we use the same input views to render 16 novel views, querying the same target viewpoint but specifying times as \(t = \left\{ \frac{i}{16} \cdot 2\pi \right\}_{i=0}^{15}\). As shown in \cref{fig:cycle}, TimeNeRF correctly produces distinct appearances for each specific time, capturing transitions that resemble the natural shifts seen throughout the day. 

\begin{figure}
\centering
  \includegraphics[scale=0.47]{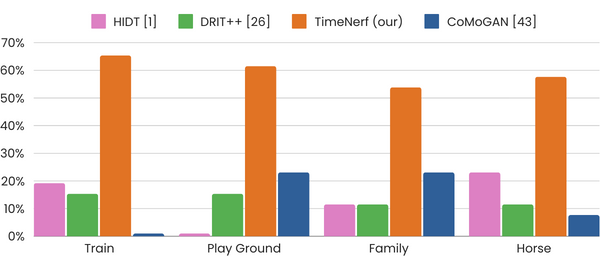}
  \caption{\textbf{Novel view synthesis across times.} We conduct a user study to ask subjects to select results that are more consistent translation between frames and accurately corresponds to the time depicted in the reference image on T\&T dataset. The number represents the percentage of preference based on 25 trial participants. }
  \label{fig:user_study}
\end{figure}

\noindent\textbf{Comparison. } To the best of our knowledge, no existing method is designed for rendering novel views across time. For comparison, we combine view synthesis algorithms with image translation models, which transform images to reflect different times of the day. We utilize our model to produce novel views. Next, DRIT++ \cite{DRIT_plus}, HiDT \cite{HiDT}, and CoMoGAN \cite{comogan} are used to transfer images across times. Note that \textbf{DRIT++ and HiDT require extra reference images for the desired time-relevance features}. Thus, we use frames from 24-hour time-lapse videos as reference images to generate images spanning an entire day. \textbf{Conversely, CoMoGAN, similar to our approach, can directly specify times}. The results in \cref{fig:nvs+t_compare} synthesize the \(playground\) scene in the T\&T dataset using 3 input views. While these methods can generate variations based on time, they sometimes yield undesirable effects, such as incorrect black patches during nighttime and color bias on the ground. Furthermore, all of these methods have the cross-view appearance inconsistency issue (geometric inconsistency), detailed in \cref{subsec:ViewConsistency}.

In comparison to previously discussed methods, our approach achieves more accurate cyclic appearance changes through consistent time translation. To validate this, we conducted a comprehensive user study. In this study, we presented participants with transformation results at eight distinct times of day: pre-dawn, dawn, mid-morning, afternoon, dusk, evening, and late night. These results, generated by our method and other referenced methods, were shown alongside a reference image corresponding to each specific time.
Participants were tasked with identifying the method that ensured consistent translation between frames and most accurately matched the time depicted in each reference image. The findings of this study, depicted in \Cref{fig:user_study}, demonstrate that our method facilitates smoother transitions across different times of the day. Please refer to the supplementary materials for more analysis.

\noindent\textbf{Generalizability.} Additionally, we evaluate our model on the T\&T dataset \cite{t&t} in \cref{fig:nvs+t_compare}, demonstrating its adaptability to different datasets. More results under varying conditions are shown in the supplementary material.

\begin{table*}
  \centering
  \resizebox{0.9\textwidth}{!}{
      \begin{tabular}{@{}l|cc|cc|cc|cc|cc|cc@{}}
        \toprule
        \multirow{3}{*}{Method}  & 
        \multicolumn{2}{c|}{Ithaca365 \cite{ithaca}} & 
        \multicolumn{10}{c}{T\&T \cite{t&t}} \\
        \cmidrule{4-13}
        \cmidrule{2-3}
           &
        \multicolumn{2}{c|}{Average} &
        \multicolumn{2}{c|}{Family} &
        \multicolumn{2}{c|}{Horse} &
        \multicolumn{2}{c|}{Playground} &
        \multicolumn{2}{c|}{Train} &
        \multicolumn{2}{c}{Average} \\
          &  
        LPIPS$\downarrow$ & RMSE$\downarrow$ & 
        LPIPS$\downarrow$ & RMSE$\downarrow$ & 
        LPIPS$\downarrow$ & RMSE$\downarrow$ & 
        LPIPS$\downarrow$ & RMSE$\downarrow$ & 
        LPIPS$\downarrow$ & RMSE$\downarrow$ & 
        LPIPS$\downarrow$ & RMSE$\downarrow$ \\
        \midrule
        DRIT++*\cite{DRIT_plus}
        & 0.098 & 0.061 
        & 0.094 & 0.076
        & 0.056 & 0.110
        & 0.084 & 0.064
        & 0.052 & 0.096
        & 0.072 & 0.086 \\
        HiDT*\cite{HiDT}
        & 0.082 & 0.065
        & 0.102 & 0.090
        & 0.069 & 0.126
        & 0.076 & 0.070
        & 0.057 & 0.115
        & 0.076 & 0.100 \\
        CoMoGAN*\cite{comogan}
        & 0.164 & 0.070 
        & 0.138 & 0.086
        & 0.088 & 0.125
        & 0.144 & 0.074
        & 0.087 & 0.088
        & 0.114 & 0.093 \\
        Ours 
        & \textbf{0.058} & \textbf{0.038} 
        & \textbf{0.066} & \textbf{0.045}
        & \textbf{0.034} & \textbf{0.064}
        & \textbf{0.053} & \textbf{0.040}
        & \textbf{0.036} & \textbf{0.058}
        & \textbf{0.047} & \textbf{0.052} \\
        \bottomrule
      \end{tabular}
    }
  \caption{\textbf{Comparison of view consistency.} We evaluate the consistency scores (LPIPS and RMSE) using 15 scenes from the Ithaca365 dataset, each with 2 target views; 4 scenes from the T\&T dataset, each with 15 pairs of target views. For each target view, we generate results at 16 different time points. Our method achieves better cross-view consistency in different datasets.}
  \label{tab:consistency}

\end{table*}

\subsection{View Consistency}
\label{subsec:ViewConsistency}
Following \cite{Huang22StylizedNeRF,learnStyle,style3D}, we employ the warped LPIPS metrics \cite{lpips} and RMSE to measure consistency across different views. The score is computed by \(E(O_v, O_{v'}) = f(O_v, W(O_{v'}), M_{v,v'})\), where \(O_v\) and \(O_{v'}\) represents the generated image \(\hat{I}_v, \hat{I}_{v'}\), and their camera parameters.
We warp the result from view \(v'\) to view \(v\) using the depth estimation in our model, which is estimated by replacing \(\mathbf{c}_i\) in \cref{eq:volumerendering} to the depths of sample points. This process is denoted by \(W(O_{v'})\). \(M_{v,v'}\) is a mask of valid pixels and \(f\) is the measurement metric, such as LPIPS or RMSE. Both quantitative and qualitative analyses (\cref{tab:consistency} and \cref{fig:viewinconsist_compare}) demonstrate that our method achieves better cross-view consistency. This improvement stems from our innovative approach of directly modeling time-relevant appearance changes within a 3D space, in contrast to the image translation approaches that operate in 2D space.

\subsection{Ablation Study}
\label{subsec:AblationStudy}

  
  


  


  

\noindent\textbf{The designed loss functions.} We study the effectiveness of our designed loss functions. The reconstruction loss is designed to ensure that the model renders views with accurate appearances when the input time coincides with the input views' capture time. Moreover, it allows the rendered views to adapt based on these inputs for different times. In \cref{fig:ablation}(a), the model without \({L}_{t_0}\) generates a road that appears blue at \(t=t_0\), although it should be gray. The delta t loss aims to ensure that the model produces unique outputs for each \(t \in [0,2\pi)\). \cref{fig:ablation}(b) shows the results obtained when training the model without \(L_{\Delta t}\). We observe that the model renders similar outputs at \(t=\frac{\pi}{2}\) and \(t=\frac{3\pi}{2}\).
\\[3.5pt]
\noindent\textbf{Two-branch network. } In \cref{fig:ablation}(c), we note that omitting the two-branch network from our time-dependent radiance field constructor (\cref{subsec:Time-DependentRadianceFieldConstructor}) leads to mapping failures within \( [0,2\pi) \). The model generates nearly black-and-white scenes at \( t=0 \) and \( t=\frac{\pi}{2} \). This limitation is due to the lack of the first branch, which serves as a template to constrain the output. By incorporating our proposed two-branch network and the designed loss functions (\cref{fig:ablation}(d)), we achieve consistent and accurate results across all time points.

\subsection{Few-Shot View Synthesis}
\label{subsec:NovelViewSynthesis}
To demonstrate TimeNeRF's capability in the traditional novel view synthesis task, we evaluate it on the Ithaca365 and LLFF datasets \cite{llff}, comparing its performance with other few-shot generalizable 3D modeling approaches. 
For a fair comparison, we train all the models on the Ithaca365 dataset.
The quantitative comparison for the Ithaca365 and LLFF datasets is presented in \cref{tab:nvs}. While TimeNeRF’s primary innovation lies in modeling the temporal dynamics of 3D scenes, this experiment reveal that our model outperform MVSNeRF and matches the capabilities of GeoNeRF in in view synthesis.

\begin{table}
  \centering
  \resizebox{0.43\textwidth}{!}{
  \begin{tabular}{@{}l|ccc|ccc@{}}
    \toprule
    \multirow{2}{*}{Method} & 
    \multicolumn{3}{c|}{Ithaca365 \cite{ithaca}} & \multicolumn{3}{c}{LLFF \cite{llff}} \\
       & PSNR$\uparrow$ & SSIM$\uparrow$ & LPIPS$\downarrow$ 
       & PSNR$\uparrow$ & SSIM$\uparrow$ & LPIPS$\downarrow$ \\
    \midrule
    MVSNeRF\cite{mvsnerf} & 25.90 & 0.703 & 0.485 & 16.58 & 0.513 & 0.503\\
    GeoNeRF\cite{GeoNeRF}  & \underline{27.46} & \underline{0.768} & \textbf{0.364} & \underline{19.62} & \textbf{0.597} & \textbf{0.415}\\
    Ours & \textbf{27.67} & \textbf{0.773} & \underline{0.367} & \textbf{19.71} & \underline{0.593} & \underline{0.418}\\
    \bottomrule
  \end{tabular}
  }
  \caption{\textbf{Novel view synthesis. }The best and the second best result are highlighted in \textbf{bold} and \underline{underline}, respectively.}
  \label{tab:nvs}
\vspace{-20pt}
\end{table}


\CCH{
\section{Conclusion and Discussion}
\label{sec:conclusion}

We propose TimeNeRF, a novel framework for view synthesis that renders views at any time from limited input views without per-scene optimization. We have effectively leveraged existing datasets to address the lack of data availability for the aforementioned task. Our model transitions smoothly across time by creating a content radiance field and transforming it into a time-dependent radiance field. In addition, our designed loss functions ensure cyclic changes and adaptive results based on inputs. Evaluations show TimeNeRF's capability to produce photorealistic views across time. 

As the initial solution in few-shot, generalizable novel view synthesis across arbitrary viewpoints and times, TimeNeRF has showcased its efficacy on various datasets. Nevertheless, significant opportunities for advancement remain. For a comprehensive 3D scene model, the dynamics of lighting angles from diverse sources—such as the sun and streetlights—and the associated constraints of object shadows relative to these light sources need to be accounted for. Additionally, achieving appearance and view consistency in both static and dynamic objects throughout varying times of the day poses a substantial challenge that could dramatically enhance user experience.
Moreover, incorporating diffusion models as a tool for data augmentation presents a promising direction to address the scarcity of available data, which could catalyze further enhancements in model performance. These considerations may underline future research on 3D modeling technologies. 
}


\begin{acks}
This work was financially supported in part (project number: 
112UA-10019) by the Co-creation Platform of the Industry Academia Innovation School, NYCU, under the framework of the National Key Fields Industry-University Cooperation and Skilled Personnel Training Act, from the Ministry of Education (MOE) and industry partners in Taiwan.  It also supported in part by the National Science and Technology Council, Taiwan, under Grant NSTC-112-2221-E-A49-089-MY3, Grant NSTC-110-2221-E-A49-066-MY3, Grant NSTC-111- 2634-F-A49-010, Grant NSTC-112-2425-H-A49-001-, and in part by the Higher Education Sprout Project of the National Yang Ming Chiao Tung University and the Ministry of Education (MOE), Taiwan. 
\end{acks}

\bibliographystyle{ACM-Reference-Format}
\bibliography{sample-base}

\clearpage
\section*{Additional Implementation Details}
\label{sec:AdditionalImplementationDetails}

\subsection{Architecture}

\subsubsection{The modifications of DRIT++\\} 
\begin{figure}
  \centering
  \includegraphics[scale=0.35]{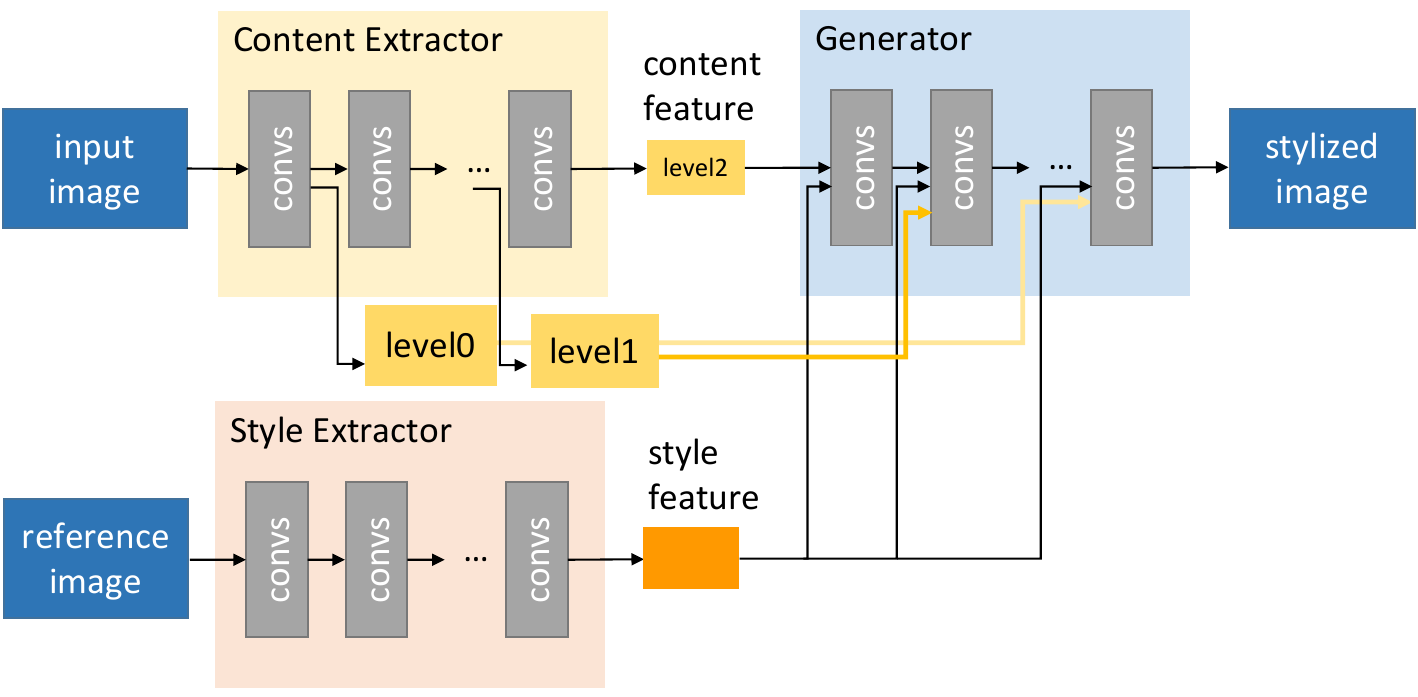}
  \caption{\textbf{DRIT++ modification.}  To improve the content extractor in DRIT++, we extract content features at three levels from the convolutional layers and merge each of them with the style feature in the generator.}
    \Description{DRIT++ modification: To improve the content extractor in DRIT++, we extract content features at three levels from the convolutional layers and merge each of them with the style feature in the generator to produce the stylized image.}
  \label{fig:dritModify}
\end{figure}

In Stage 1, we utilize DRIT++ \cite{DRIT_plus} for feature disentanglement. 
DRIT++ \cite{DRIT_plus} introduces a disentangled representation framework designed to transfer source content into a specified style. It achieves this by mapping images into two distinct spaces: a domain-invariant content space that captures shared information across domains, and a domain-specific attribute space. During synthesis, the model uses the encoded content features from a given input and attribute vectors sampled from the attribute space to generate diverse and high-quality outputs that reflect the desired style.
As described in the main paper, we extract content features at three different levels from different convolutional layers of the content extractor, in contrast to the original DRIT++, which relies only on the final layer output of the content extractor. These multi-level content features capture different semantic information. We then merge each level of content features with the style feature to generate stylized images, as illustrated in \cref{fig:dritModify}. Utilizing features from multiple layers in the generation process encourages the model to leverage more content features and enhances the model's capabilities for stylized image generation.

\subsubsection{Details of the implicit scene network\\} The specific implementation details of \(H(\cdot)\), mentioned in Sec. 3.4 of the main paper, are adapted from GeoNeRF \cite{GeoNeRF}. Following GeoNeRF, we employ Multi-Head Self-Attention (MHSA) \cite{transformer} and full-connected layers to aggregate information of different input views. Below, we explain the procedure:

First, to facilitate the exchange of information between different views, features are aggregated via MHSA layers \cite{transformer} by the following equations.
\begin{equation} \label{eq:MHA}
    \tilde{\sigma}^\mathbf{x}, \{\tilde{w}_i^\mathbf{x}\}_{i=1}^N = \text{MHSA}(h(cf^\mathbf{x}), \{g_i^\mathbf{x}\}_{i=1}^N).
\end{equation}
\begin{equation}
    h(z) = \text{FC}(\text{mean}(z) || \text{var}(z)).
\end{equation}
Here, \(cf^\mathbf{x}\) represents \(\{cf_i^\mathbf{x}\}_{i=1}^{N}\), the content features of \(N\) input views. \(g_i^\mathbf{x}\) is the interpolation of the geometry feature at \(\textbf{x}\) from the input view \(i\). \(||\) is the concatenation and FC refers to fully-connected layers. MHSA layers combine features from different views, producing enhanced features \(\tilde{\sigma}\) and \(\{\tilde{w}_i\}_{i=1}^N\).

Then, an auto-encoder (AE) network is applied to the features \(\{\tilde{\sigma}_p\}_{p=1}^M\) of all the sample points \(\{\textbf{x}_p\}_{p=1}^M\) along the ray \(\mathbf{r}\) for facilitating information exchange along the ray and achieve the updated density features \(\{\sigma'_p\}_{p=1}^M\) of the ${M}$ points by 
\begin{equation}
    \{\sigma'_p\}_{p=1}^M=\text{AE}\left(\{\tilde{\sigma}_p\}_{p=1}^M\right),
\end{equation}
where \(\tilde{\sigma}_p\) is obtained from \cref{eq:MHA} for the sample point \(\mathbf{x}_p\) in the ray \(\mathbf{r}\).
After enhancing the geometry information, an MLP is used to predict the final density of each 3D sample point by
\begin{equation}
    \sigma = \text{MLP}_{\sigma}(\sigma').
\end{equation}

\begin{figure}
  \centering
  \includegraphics[scale=0.35]{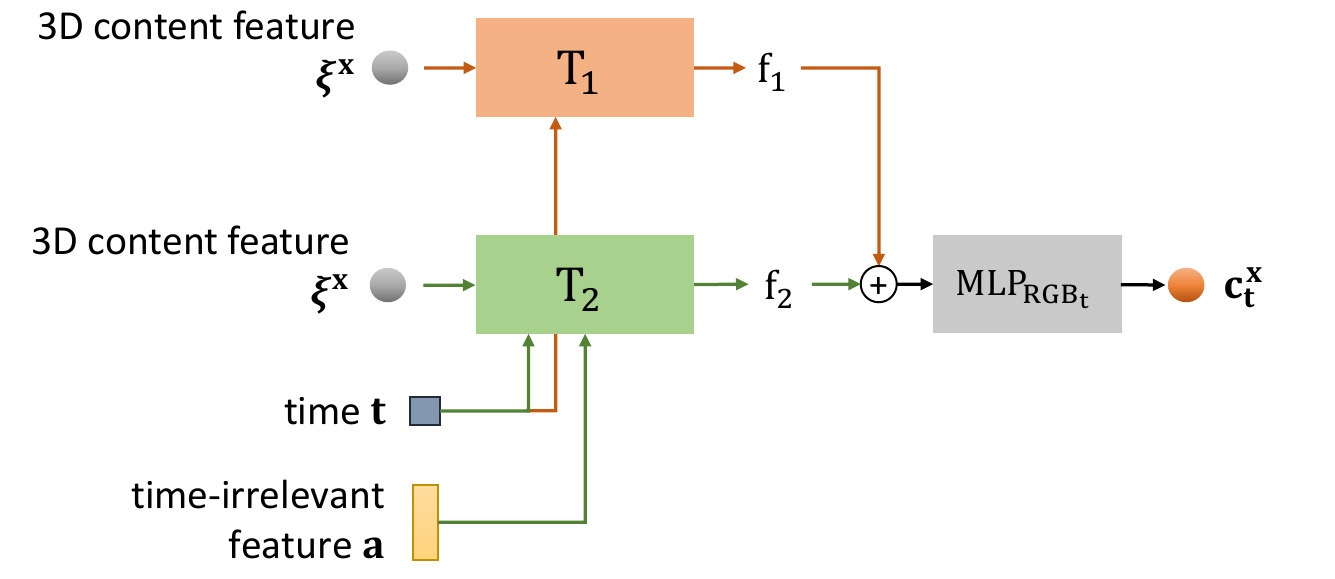}
  \caption{\textbf{Time-Dependent Radiance Field Constructor.} We have developed a two-branch network for the Time-Dependent Radiance Field Constructor to adequately model the temporal variations.}
  \Description{}
  \label{fig:T()}
\end{figure}

\begin{figure*}
\vspace{2cm}
  \centering
  \includegraphics[scale=0.3]{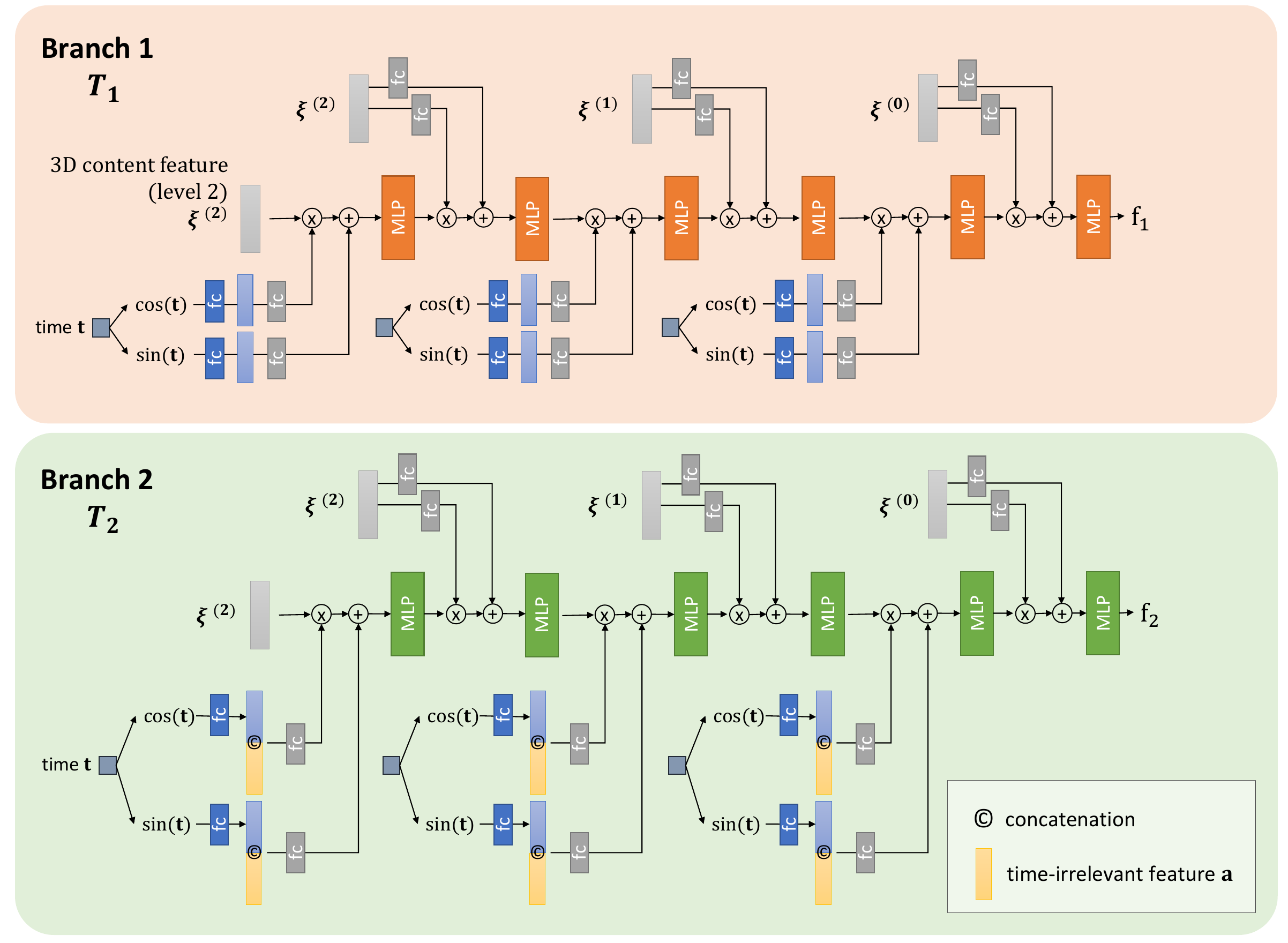}
  \caption{\textbf{Details of \(\text{T}_1\) and \(\text{T}_2\).} The figure illustrates the architecture design of \(T_1\) and \(T_2\) used in the time-dependent radiance field constructor (\cref{fig:T()}).}
  \label{fig:T1T2}
\end{figure*}

\subsubsection{Details of the time-dependent radiance field constructor\\}

As discussed in Sec. 3.5 of the main paper, our time-dependent radiance field constructor, \(T\), uses a two-branch network to convert the content radiance field into the time-dependent radiance field.
The architecture design, as shown in \cref{fig:T()} including branches \(T_1\) and \(T_2\) along with an MLP for color decoding, ensures our constructor adequately models the temporal variations. Specifically, the first branch \(T_1\) combines 3D content feature \(\xi\) with time \(t\),
to serve as the template for the change over time. The second branch \(T_2\) integrates 3D content feature \(\xi\) with both time \(t\) and time-irrelevant feature \(a\) to further tune the color according to time-irrelevant features \(a\). Finally, the output features of \(T_1\) and \(T_2\) are then summed and passed through an MLP to generate the time-dependent color \(\textbf{c}_t\).

The specifics of \(T_1\) and \(T_2\) are shown in \cref{fig:T1T2}. We use content features at three levels \(\{\xi^{(l)}\}_{l=0}^2\) and integrate them from the high-level features (i.e, capturing more global information) to the low-level features (i.e., focusing on local details) along with time-related style information.

\subsection{Training an inference extra details}

\subsubsection{Stage 1 training\\} 
\begin{figure}
  \centering
  
  \begin{subfigure}{\linewidth}
  \centering
    \begin{tabular}{@{}c@{\hspace{1mm}}c@{\hspace{1mm}}c@{}}
      {\includegraphics[scale=0.42]{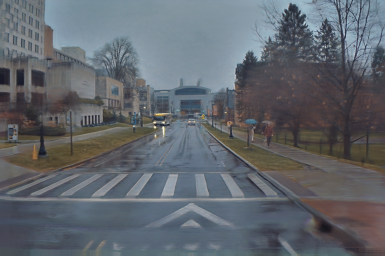}} & 
      {\includegraphics[scale=0.42]{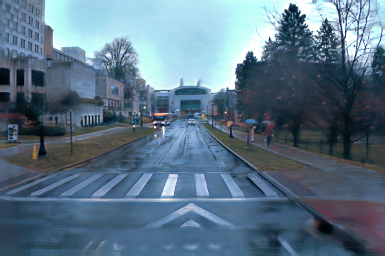}} & 
      {\includegraphics[scale=0.42]{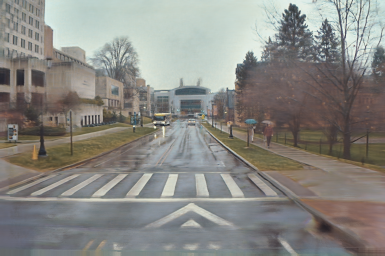}} \\
      \parbox[t]{2.2cm}{\centering Input image} & 
      \parbox[t]{2.2cm}{\centering (a) CoMoGAN } &
      \parbox[t]{2.2cm}{\centering (b) TimeNeRF }
    \end{tabular}
  \end{subfigure}

  \caption{\textbf{CoMoGAN and TimeNeRF under a rainy day.} (a) is the stylized result from CoMoGAN, which converts the rainy scene to the daytime condition. However, it generates a color-biased scene (i.e., the sky). (b) In contrast, TimeNeRF is able to translate images according to different weather conditions without a color bias. More examples can be found in \cref{fig:diverseWeather}.}
  \label{fig:comoganRain}
\end{figure}

To manipulate the time, one possible way is to disentangle time-related information from images in Stage 1. However, doing so without supervision across diverse weather conditions is challenging. Weather factors such as lighting and shadows can interfere with accurate time information extraction. While training the model on images from a single weather condition might simplify the task, it risks overfitting to that specific weather condition. For example, CoMoGAN \cite{comogan}, which has the capability of continuous time translation, is trained on a dataset of images captured at different times but only on sunny days. As shown in \cref{fig:comoganRain}, when an input image captured on a rainy day is transferred to daytime, it produces a color-biased sky, which is unreasonable.

To avoid this, we first extract style features from images in Stage 1 that encompass all environmental change factors, including both time and weather (time-irrelevant) information. In Stage 2, we then disentangle these factors. By training on a diverse set of weather conditions, our training approach prevents the model from overfitting to a specific weather condition and makes it easier to separate time and weather information from the style feature.

\subsubsection{The inputs of the factor extraction module\\} 

As mentioned in the main paper, we hypothesize that the extracted style encompasses both time and weather information and propose to disentangle these two factors utilizing the factor extraction module. As detailed in Sec. 3.5 of the main paper, we employ two Multi-Layer Perceptrons (MLPs) to predict time \(t\) and extract time-irrelevant information (e.g. weather) denoted as \(a\). In the following, we explain our reasoning for extracting \(t\) from a reference image and \(a\) from an input image during training.

The objective of our model is to learn the time transitions occurring throughout the day while preserving the weather conditions present in the input data. To this end, we extract the time-irrelevant feature \(a\) from the input view to preserve the weather conditions and remove the time-relevant part from the style feature. For learning time transitions, we utilize reference images captured at various times to provide the model with additional time information during training. Specifically, we extract time-related factors \(t\) from these reference images and learn to map the style features of reference images into \([0,2\pi)\). In the testing phase, our model is capable of simulating time transitions by specifying the time \(t\) directly without referring to any image, making our method different from the reference-based style transfer methods like DRIT++~\cite{DRIT_plus} and HiDT~\cite{HiDT}.

\subsubsection{Model size, training, and inference time\\} The model consists of 58.3 million parameters. To ensure generalizability, we utilize MVSnet~[62] as our pretrained model for feature extraction. The training process spans 250,000 iterations and requires approximately 2 days on a single NVIDIA RTX 3090. During the testing phase, synthesizing novel views over time takes around 30 seconds.



\section{Additional Experimental Results}
\label{sec:AdditionalExperiments}

\subsection{Comparison with few-shot IN2N}

\begin{figure}[htbp]
  \centering
  \includegraphics[scale=0.1]{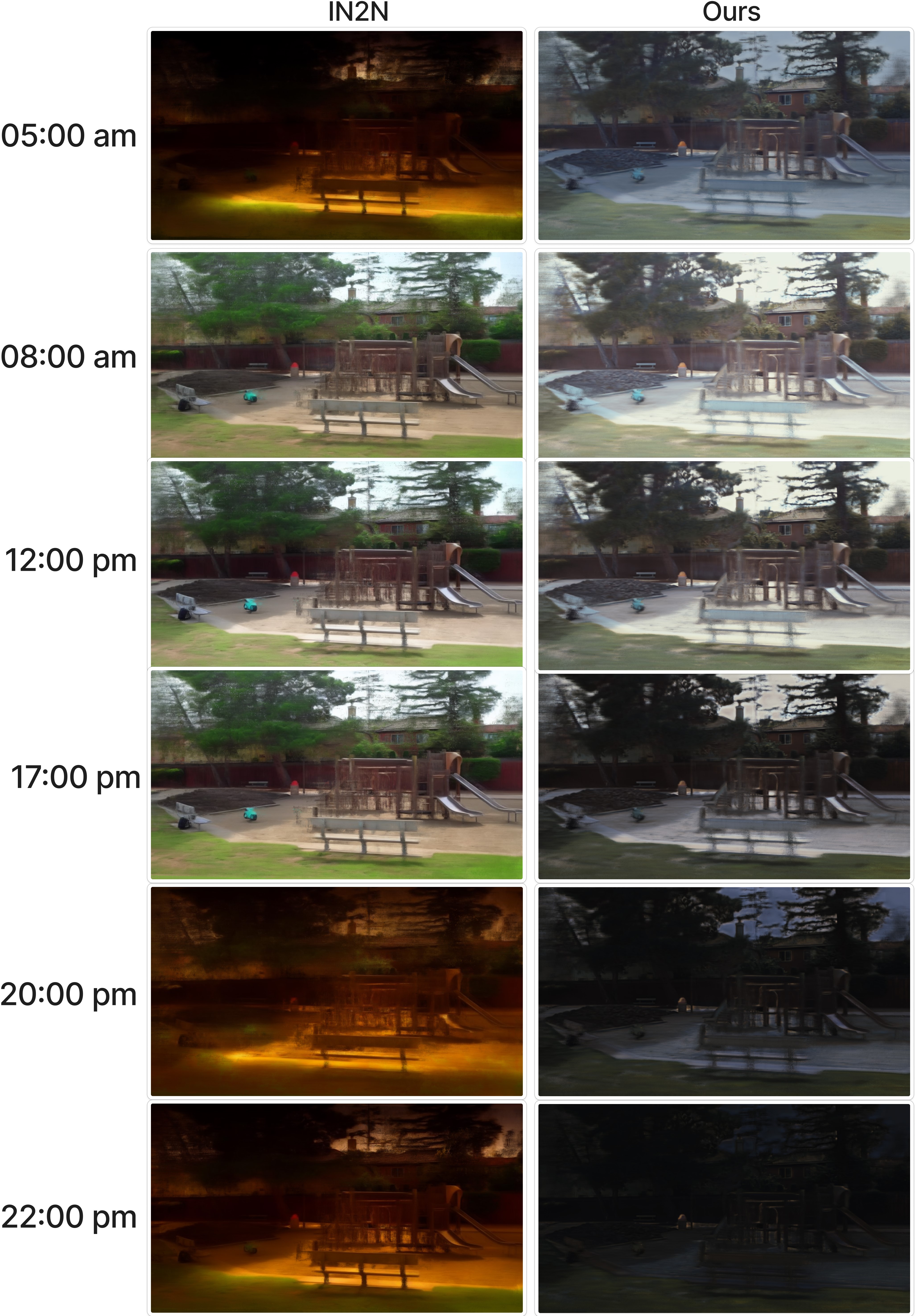}
  \caption{Time consistency comparison}
  \Description{}
  \label{fig:n2n_timenerf}
\end{figure}

For a fair few-shot comparison, we integrate ``GeoNeRF'' with ``InstructPix2Pix'' to create a 3D NeRF over time using three reference views, termed few-shot InstructNeRF2NeRF (IN2N). During the experiment, the text prompts follow the format \textit{``change image time to \{times of day\}''}. The comparisons, evaluated on the T\&T dataset, are shown in Fig. \ref{fig:n2n_timenerf} and Table \ref{tab:style_consitancy}.

\textbf{Inconsistent across timesteps.} Compared to TimeNeRF, the novel views generated by IN2N (Fig. \ref{fig:n2n_timenerf}) exhibit time inconsistency, as they typically produce only two modes (day and night) without successive appearance changes. Moreover, Table \ref{tab:style_consitancy} indicates that TimeNeRF demonstrates superior time-style consistency. Accordingly, TimeNeRF shows better abilities to capture continuous appearance evolution over time compared to IN2N, which relies on text prompts to control time properties.

\textbf{Quality issue of novel views.} In a few-shot setting with only three reference image views, the quality of novel views is naturally reduced (i.e., more shots would significantly improve quality). However, as shown in Table 2 of section 4.5, our method outperforms other SOTA NeRF-based methods in terms of PSNR and SSIM. Also, Fig. \ref{fig:n2n_timenerf} shows that IN2N produces blurry novel views. Besides, IN2N faces view inconsistency issues due to independent style transfer for each reference view. In contrast, TimeNeRF does not have the view inconsistency issue (section 4.4).

\vspace{-8pt}
\begin{table}[htbp]
  \centering
  \begin{tabular}{@{}l|cc@{}}
    \toprule
    \multirow{1}{*}{Time of Day}
       & IN2N  & Ours   \\
    \midrule
    Day & 9.0664 & \textbf{8.8019}\\
    Dusk/Dawn  & 9.1345 & \textbf{8.7307}\\
    Night  & 10.6849 & \textbf{10.4031}\\
    \bottomrule
  \end{tabular}
  \caption{Style Consistency Analysis. We use FID to assess style consistency between novel and reference images across time. Lower FID values indicate higher time-style similarity.}
  \label{tab:style_consitancy}
\end{table}

\subsection{Ablation study}

\begin{figure}
\vspace{2cm}
  \centering
  
  \begin{subfigure}{\linewidth}
    \centering
    \begin{tabular}{@{}cc@{}}

      Input image & Reference image \\ 
      \includegraphics[scale=0.9]{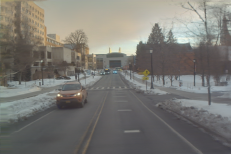} & 
      \includegraphics[scale=0.9]{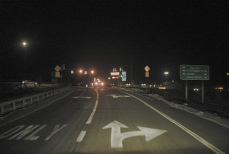}  \\

      \midrule

      \includegraphics[scale=0.9]{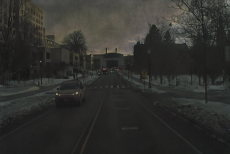} & 
      \includegraphics[scale=0.9]{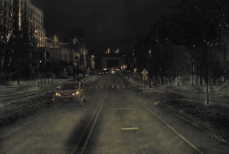} \\
      (a) Original DRIT++ & (b) Modification \\
      
    \end{tabular}
  \end{subfigure}

  \caption{\textbf{Ablation study on DRIT++ modification.} According to the reference image, we transformed the input image into a stylized night-time image using (a) the original DRIT++ and (b) our modified model.}
  \label{fig:ablationDrit}
\end{figure}

\noindent\textbf{DRIT++ modification. } As shown in \cref{fig:ablationDrit}(a), some residual light remains in the sky within the style-transferred image generated by the original DRIT++~\cite{DRIT_plus}. This is likely because the content feature in DRIT++ still contains some daytime information. In contrast, our modified DRIT++, considering features in three levels, produces a more accurate transferred result (\cref{fig:ablationDrit}(b)).

\subsection{Under diverse weather conditions}
\begin{figure*}
  \centering
  
    \hspace{-3.4cm}
    \begin{tabular}{@{}c@{\hspace{1.5mm}}c@{}}
      
      \parbox[t]{3cm}{\centering (a) Rainy day} & \raisebox{-0.9\height}{\includegraphics[scale=0.5]{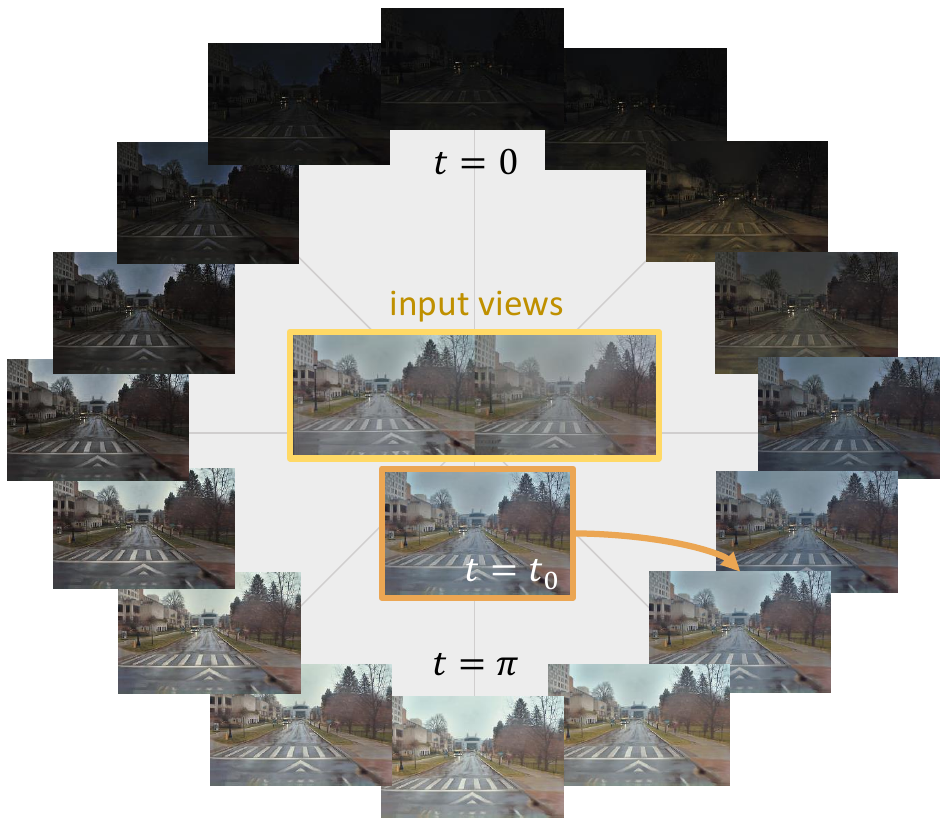}} \\
      \parbox[t]{3cm}{\centering (b) Snowy day} & \raisebox{-0.9\height}{\includegraphics[scale=0.5]{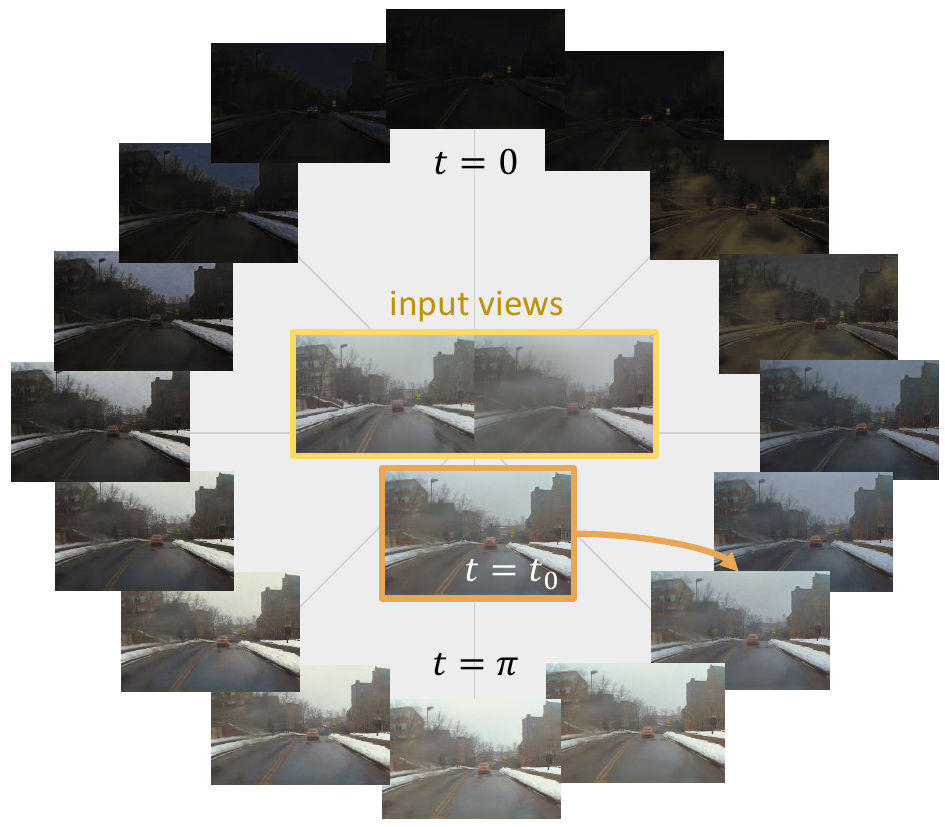}} \\
      \parbox[t]{3cm}{\centering (c) Cloudy day} & \raisebox{-0.9\height}{\includegraphics[scale=0.5]{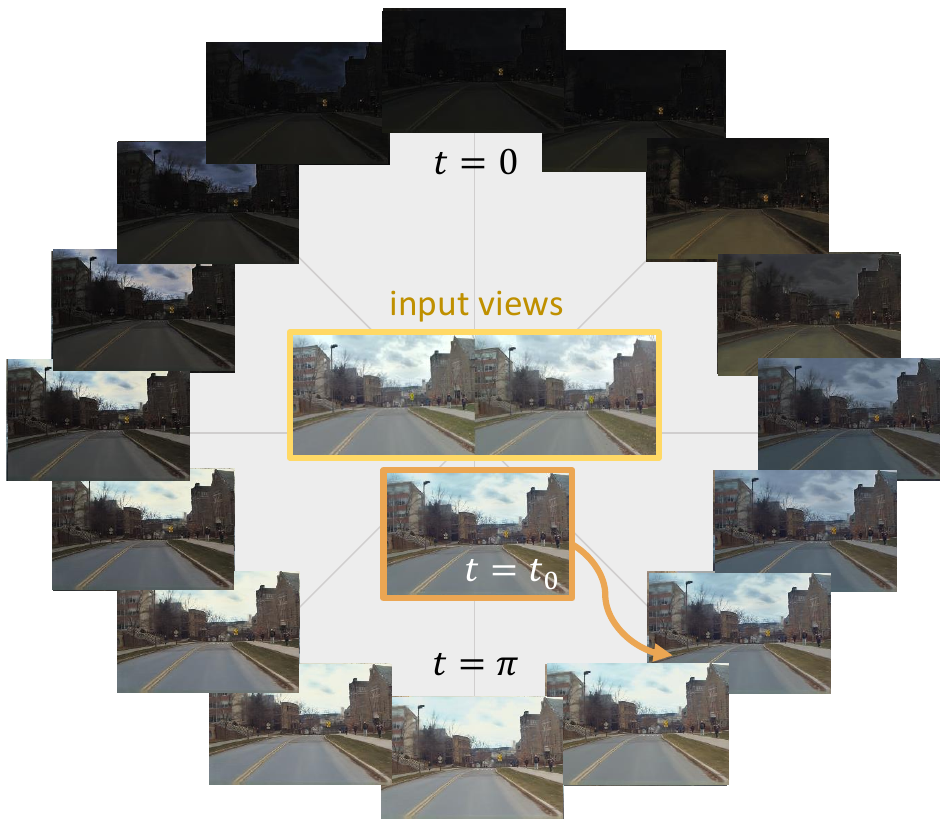}}
      
    \end{tabular}

  \caption{\textbf{Synthesis under diverse weathers.} We show the synthesis results at 16 different time points. The input images are captured on a (a) rainy, (b) snowy, and (c) cloudy day. For each scene, two input images are utilized for 3D reconstruction. The images in the yellow box represent the two input views of a test scene. The images around the circle are novel views at different times. The image in the orange box is synthesized for the time of input views \(t_0\).}
  \Description{}
  \label{fig:diverseWeather}
\end{figure*}

Besides the results shown in Fig. 3 of the main paper, to evaluate our model's performance of novel view synthesis across times under varied conditions, we test TimeNeRF with views captured in diverse weather scenarios (i.e.,  rainy, snowy, and cloudy days). The synthesized results with texture consistency and temporal smoothness shown in \cref{fig:diverseWeather} demonstrate the robustness of TimeNeRF to diverse weather conditions.

\subsection{Analysis of Possible Alternative Approaches and Their Weaknesses}

Alternative approaches that aim to achieve a similar system goal as the proposed TimeNeRF can be considered. However, these alternatives have certain weaknesses that make them unsuitable solutions. Below, we detail these issues.

\textbf{Novel view synthesis, then style transfer: }
Synthesizing novel views and subsequently transferring styles can lead to view/geometry inconsistency issues, as discussed in Section 4.4 of the main paper. This issue arises because the style transfer model lacks awareness of 3D geometry, which results in the introduction of unrealistic effects in the scene without considering its underlying 3D structure. Consequently, these methods may struggle with artifacts due to their reliance on 2D information.

\textbf{Style transfer, then novel view synthesis: } On the other hand, applying style transfer to input images before reconstructing a 3D scene and synthesizing novel views may prove ineffective. Given that style transfer operates in 2D space, it often merges inconsistent styles captured from different viewpoints into the 3D scenes. This inconsistency can lead to inaccuracies in NeRF geometry, resulting in imprecise synthesized novel views. Some examples of these inaccuracies are illustrated in \cref{fig:style_novelview}.

\subsection{Additional Qualitative and Quantitative Studies}

\subsubsection{Qualitative results.\\}

\begin{figure*}
\vspace{1cm}
  \centering

  \begin{subfigure}{\linewidth}
      \raisebox{0.\height}{\includegraphics[scale=0.473]{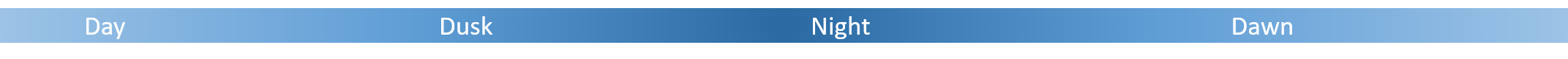}}
  \end{subfigure}
  
  \begin{subfigure}{\linewidth}
      \raisebox{-0.5\height}{\includegraphics[scale=0.46]{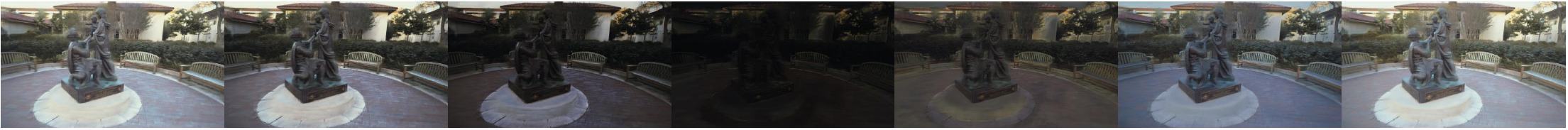}}
  \end{subfigure}

  \begin{subfigure}{\linewidth}
      \raisebox{-0.5\height}{\includegraphics[scale=0.46]{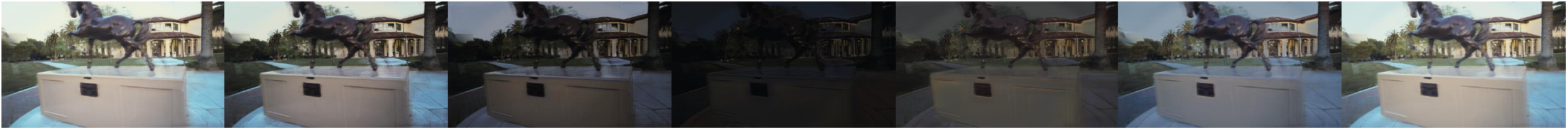}}
  \end{subfigure}

  \begin{subfigure}{\linewidth}
      \raisebox{-0.5\height}{\includegraphics[scale=0.46]{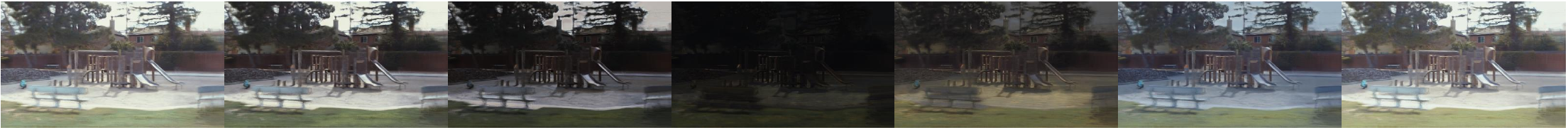}}
  \end{subfigure}

  \begin{subfigure}{\linewidth}
      \raisebox{-0.5\height}{\includegraphics[scale=0.46]{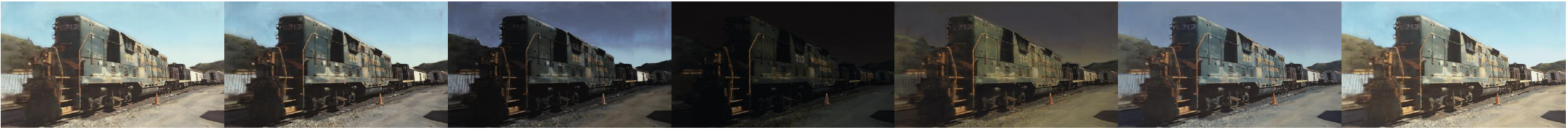}}
  \end{subfigure}
  
  \caption{\textbf{Qualitative results on the T\&T~\cite{t&t} dataset.} We generate novel views at 7 different times to show the cyclic changes of a day. For each scene, 3 input images are utilized for 3D reconstruction in this experiment.}
  \label{fig:TTresult}
\end{figure*}

\begin{figure*}
\vspace{1cm}
  \centering
  
    \begin{tabular}{@{}cc@{}}
      
       & \makebox[2cm]{Novel view GT} \hspace{1.5cm} \makebox[2cm]{MVSNeRF \cite{mvsnerf}} \hspace{1.5cm} \makebox[2cm]{GeoNeRF \cite{GeoNeRF}} \hspace{1.5cm} \makebox[2cm]{Ours} \\
       
      \parbox[t]{1.5cm}{\centering \large Ithaca365 \cite{ithaca}} & \raisebox{-0.5\height}{\includegraphics[scale=0.51]{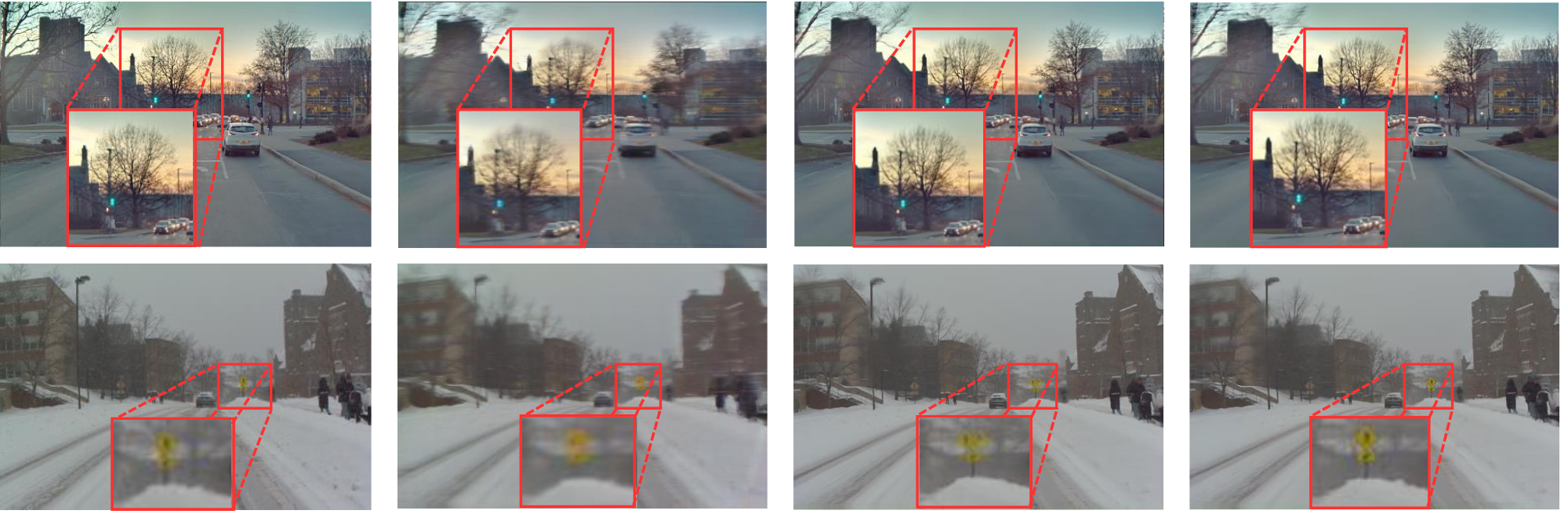}} \\
      \midrule
      \parbox[t]{1.5cm}{\centering \large LLFF \cite{llff}} & \raisebox{-0.5\height}{\includegraphics[scale=0.51]{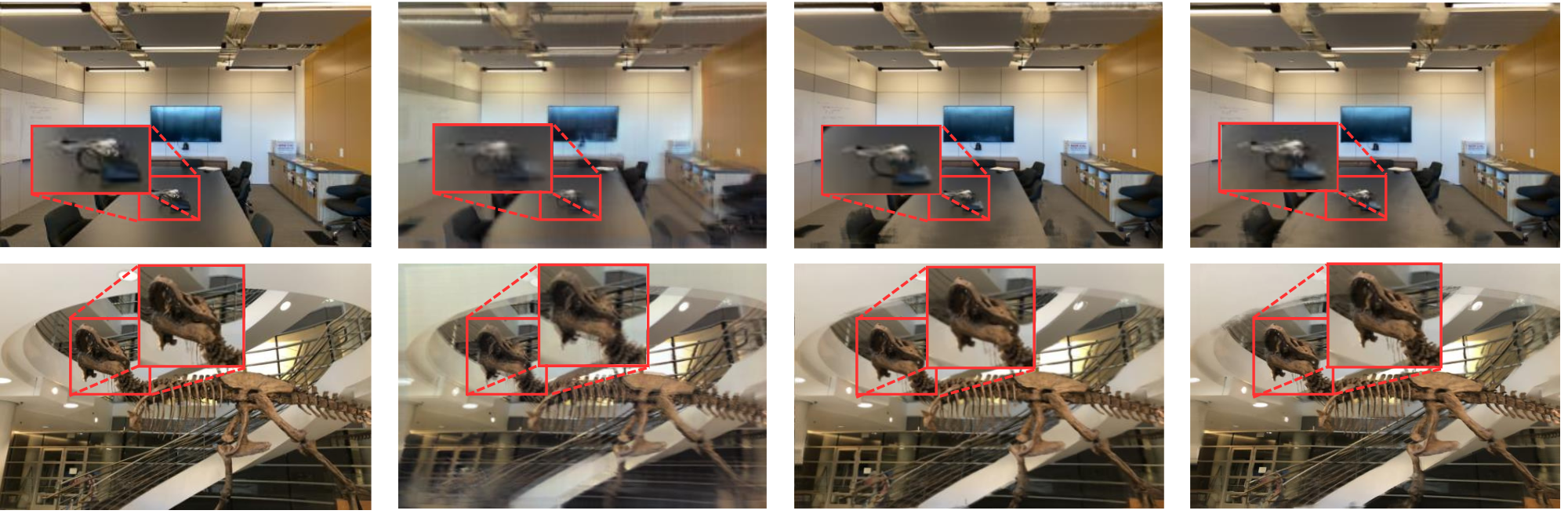}} \\
      
    \end{tabular}

  \caption{\textbf{Qualitative results of pure view synthesis.} We show the view synthesis results from MVSNeRF, GeoNeRF, and our model on the LLFF dataset and the ithaca365 dataset. Compared to these SOTA methods, TimeNeRF is able to produce clearer details.}
  \label{fig:nvs}
\end{figure*}

In \cref{fig:TTresult}, we showcase additional synthesized novel views for the Family, Horse, Playground, and Train scenes from the T\&T dataset \cite{t&t}. These results demonstrate our model's ability to smoothly generate novel views over time in diverse scenarios. Additionally, we present more qualitative results from the LLFF dataset \cite{llff} and the Ithaca365 dataset \cite{ithaca} in \cref{fig:nvs}. This provides a comparative analysis against state-of-the-art (SOTA) NeRF-based methods focused purely on novel view synthesis without temporal variations. Compared to these SOTA methods, TimeNeRF is able to produce clearer details.


\subsubsection{Quantitative results.\\}

To further assess the quality of synthesized images across various times, we conduct two additional analyses and comparisons. First, we evaluate color consistency across different periods by calculating the mean histogram correlation for the Y, Cb, and Cr color channels over time. This analysis helps determine how well color properties are maintained throughout different phases of the day. Second, we measure the style consistency between the synthesized images and reference images to measure the overall style coherence of our image synthesis. Detailed explanations of both methodologies are provided below.

\begin{table}
  \centering
  \begin{tabular}{@{}l|cc|cc@{}}
    \toprule
    \multirow{2}{*}{Method} & 
    \multicolumn{2}{c|}{Ithaca365 \cite{ithaca}} & \multicolumn{2}{c}{T\&T \cite{t&t}} \\
       & Cb & Cr  
       & Cb & Cr \\
    \midrule
    DRIT++\cite{DRIT_plus} & 0.409 & 0.436 & 0.392 & 0.461\\
    HiDT\cite{HiDT}  & 0.315 & 0.304 & 0.302 & 0.294\\
    CoMoGAN\cite{comogan} & 0.398 & 0.544 & 0.428 & 0.612\\
    Ours & \textbf{0.702} & \textbf{0.722} & \textbf{0.614} & \textbf{0.769}\\
    \bottomrule
  \end{tabular}
  \caption{Analyzing the Mean Histogram Correlation of the Cb and Cr Color Channels Over Time. This table presents the calculated mean histogram correlation of the Cb and Cr channels, which reflects color consistency over different periods. Color consistency should be maintained across varying time intervals. Therefore, higher mean correlation values for the Cb and Cr channels are anticipated}
  \label{tab:analysisCbCr}
\end{table}

\begin{table}
  \centering
  \begin{tabular}{@{}l|cc|cc@{}}
    \toprule
    \multirow{2}{*}{Method} & 
    \multicolumn{2}{c|}{Ithaca\cite{ithaca}} & \multicolumn{2}{c}{T\&T \cite{t&t}} \\
       & CoMoGAN  & Ours  
       & CoMoGAN & Ours \\
    \midrule
    Day & 9.2933 & \textbf{8.9319} & 9.0664 & \textbf{8.8019}\\
    Dusk/Dawn  & 9.5158 & \textbf{9.3940} & 9.1345 & \textbf{8.7307}\\
    Night & 10.4778 & \textbf{10.3871} & 10.6849 & \textbf{10.4031}\\
    \bottomrule
  \end{tabular}
  \caption{\textbf{Style Consistency Analysis.} In this table, we calculate Fréchet Inception Distance (FID) between the style features of the synthesized image set and the reference set to evaluate the level of style consistency across different time periods. Smaller distance values indicate that the corresponding style is more similar to the style of the reference image at the time period. Our TimeNeRF demonstrates better style consistency compared to CoMoGAN \cite{comogan}.}
  \label{tab:feature_base}
\end{table}

\begin{figure*}
  \centering
  
  \begin{tabular}{@{}c@{\hspace{0.1em}}c@{\hspace{0.1em}}c@{\hspace{0.1em}}c@{\hspace{0.1em}}@{}}
    \small   & 
    \small Y & 
    \small Cb & 
    \small Cr \\

    \raisebox{-0.5\height}{\small Ithaca\cite{ithaca}} &
    \raisebox{-0.5\height}{\includegraphics[scale=0.35]{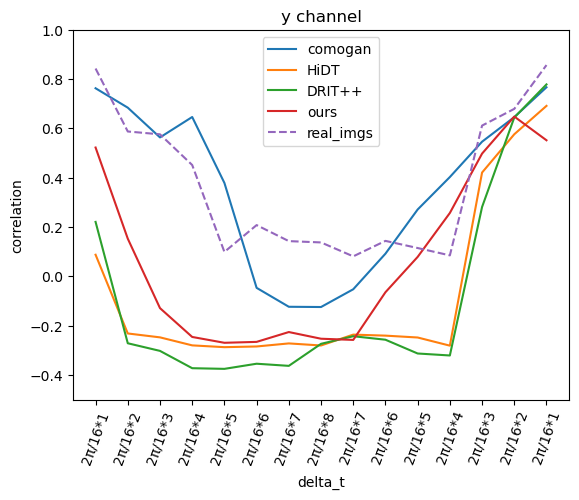}} &
    \raisebox{-0.5\height}{\includegraphics[scale=0.35]{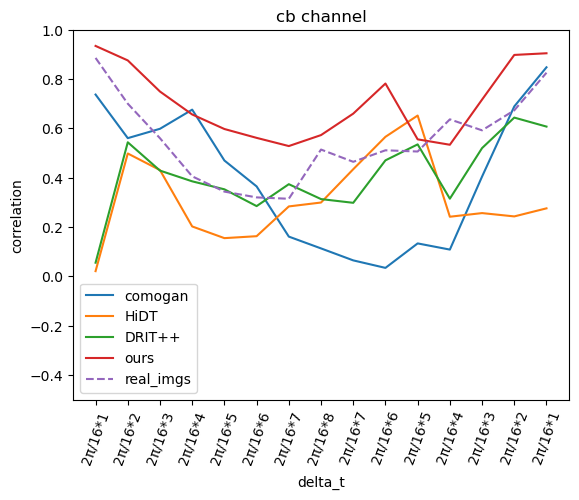}} & 
    \raisebox{-0.5\height}{\includegraphics[scale=0.35]{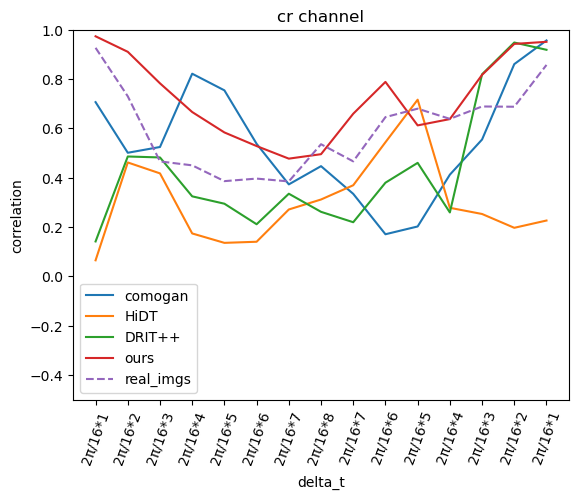}} \\

    \raisebox{-0.5\height}{\small T\&T\cite{t&t}} &
    \raisebox{-0.5\height}{\includegraphics[scale=0.35]{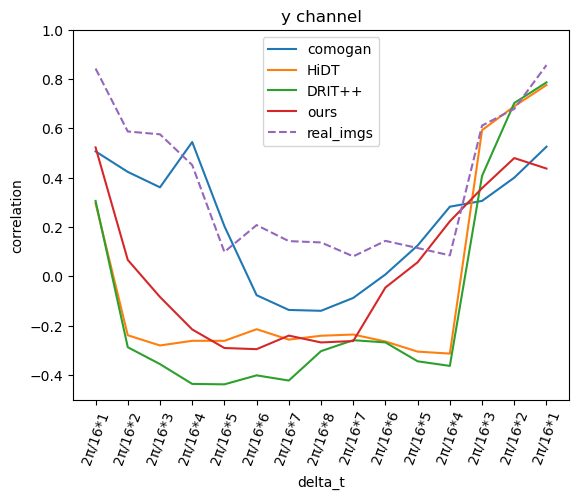}} &
    \raisebox{-0.5\height}{\includegraphics[scale=0.35]{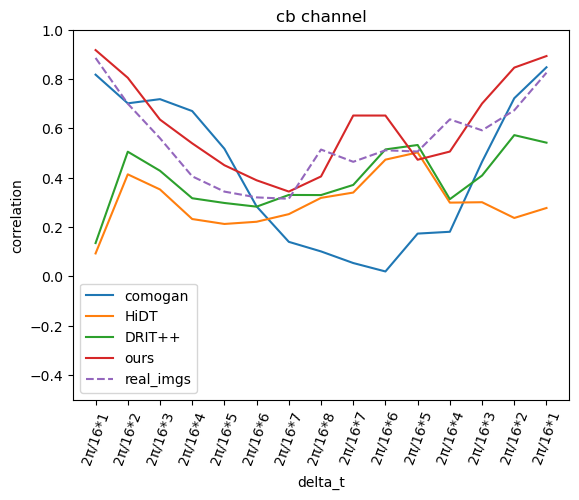}} & 
    \raisebox{-0.5\height}{\includegraphics[scale=0.35]{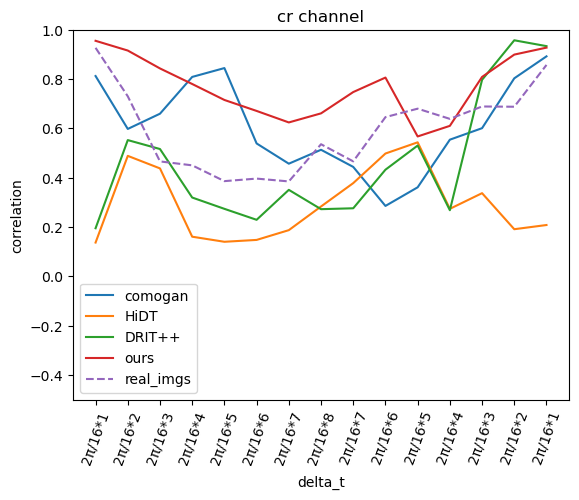}} \\
  \end{tabular}
  
  \caption{\textbf{the mean histogram correlations in the Y, Cb, and Cr channels.} The x-axis represents time differences, while the y-axis indicates correlation values. ``real\_imgs'' correspond to frames extracted from multiple 24-hour videos. Ideally, the Y channel's correlation should decrease as the time difference increases, whereas the Cb and Cr channels should exhibit relatively stable correlations compared to the Y channel.}

  \label{fig:colorHistogram}
\end{figure*}

\begin{figure*}
\vspace{1cm}
  \centering
  
    \begin{tabular}{@{}cc@{}}
      

       & \makebox[2cm]{Input view 1} \hspace{1.8cm} \makebox[2cm]{Input view 2} \hspace{1.8cm} \makebox[2cm]{Input view 3} \hspace{1.8cm} \makebox[2cm]{Synthesized Novel view} \\
       
      \parbox[t]{1.5cm} & \raisebox{-0.5\height}{\includegraphics[scale=0.05]{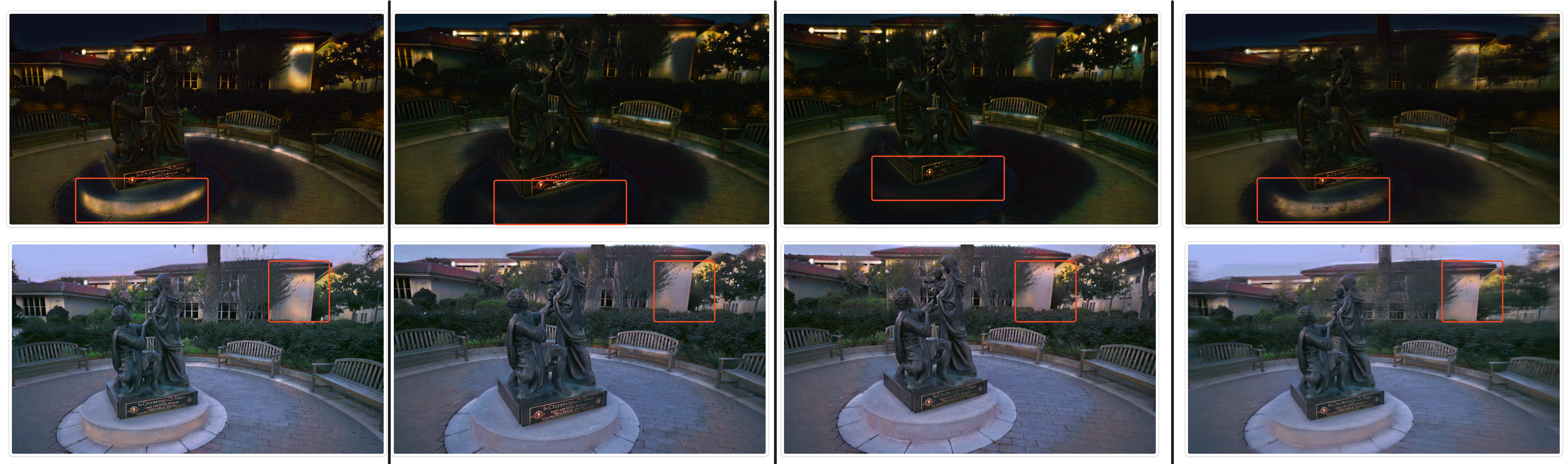}}

    \end{tabular}

  \caption{\textbf{Style transfer, then novel view synthesis. We first transfer the style of the input views using CoMoGAN \cite{comogan}, and then use these style-transferred input views to synthesize a novel view through GeoNerf \cite{GeoNeRF}. The red boxes in the images highlight regions where there are inconsistencies between the input views and the synthesized novel views produced by this alternative approach.}}
  \label{fig:style_novelview}
\end{figure*}

\textbf{Color Consistency Analysis Using YCbCr Color Space.} To demonstrate that our method provides better color consistency over time and produces fewer color biases, we analyze the mean histogram correlations in the Y, Cb, and Cr channels over time, as shown in \cref{fig:colorHistogram}. Here, we measure the correlation between the color histogram at a specific time (daytime) and those at other times. Ideally, the Y channel should show a decrease in correlation as the time difference increases, reflecting changes in illumination. In contrast, the correlations in the Cb and Cr channels should remain more consistent. Our results indicate that all methods produce a ``U'' shaped curve in the Y channel, which mirrors the real-world relationship between illumination and time difference. However, our method demonstrates greater consistency in the Cb and Cr channels. We quantify this by calculating mean correlations, with results detailed in \cref{tab:analysisCbCr}. Notably, our method achieves higher mean correlations in both the Cb and Cr channels compared to previous methods. This suggests that the lower mean correlations observed in these state-of-the-art methods may result from artifacts in the generated images. 

\textbf{Style Consistency Analysis.}
In this section, we evaluate the quality of image synthesis by measuring the style consistency between the synthesized images and the reference images. The style extractor module from DRIT++ \cite{DRIT_plus} is employed to extract style features from an image. As outlined in section 4.3 of the main paper, we synthesize novel views at three different times of the day: day, dusk/dawn, and night. For each time period, novel images are generated using TimeNeRF, and their style features are subsequently extracted. We then measure the Fréchet Inception Distance (FID) between the style features of the synthesized image set and the reference set to evaluate the level of style consistency across different time periods. The results are presented in \cref{tab:feature_base}. We assess our method using the Itheca365 \cite{ithaca} and T\&T datasets \cite{t&t}. Our proposed TimeNeRF method demonstrates better style consistency compared to the state-of-the-art CoMoGAN \cite{comogan}, showcasing its efficacy in synthesizing visually consistent images over time.



\section{The implementation code}
\noindent We are prepared to share the source code for implementation. However, due to limitations on the submission file size exceeding the model's weight, we are unable to provide the pretrained model at this time. If there are any requests regarding pretrained models for cross-checking purposes, please feel free to reach out to us

    
    



\bibliographystyle{ACM-Reference-Format}
\bibliography{sample-base}


\begin{thebibliography}{65}


\ifx \showCODEN    \undefined \def \showCODEN     #1{\unskip}     \fi
\ifx \showDOI      \undefined \def \showDOI       #1{#1}\fi
\ifx \showISBNx    \undefined \def \showISBNx     #1{\unskip}     \fi
\ifx \showISBNxiii \undefined \def \showISBNxiii  #1{\unskip}     \fi
\ifx \showISSN     \undefined \def \showISSN      #1{\unskip}     \fi
\ifx \showLCCN     \undefined \def \showLCCN      #1{\unskip}     \fi
\ifx \shownote     \undefined \def \shownote      #1{#1}          \fi
\ifx \showarticletitle \undefined \def \showarticletitle #1{#1}   \fi
\ifx \showURL      \undefined \def \showURL       {\relax}        \fi
\providecommand\bibfield[2]{#2}
\providecommand\bibinfo[2]{#2}
\providecommand\natexlab[1]{#1}
\providecommand\showeprint[2][]{arXiv:#2}

\bibitem[Anokhin et~al\mbox{.}(2020)]%
        {HiDT}
\bibfield{author}{\bibinfo{person}{Ivan Anokhin}, \bibinfo{person}{Pavel Solovev}, \bibinfo{person}{Denis Korzhenkov}, \bibinfo{person}{Alexey Kharlamov}, \bibinfo{person}{Taras Khakhulin}, \bibinfo{person}{Alexey Silvestrov}, \bibinfo{person}{Sergey Nikolenko}, \bibinfo{person}{Victor Lempitsky}, {and} \bibinfo{person}{Gleb Sterkin}.} \bibinfo{year}{2020}\natexlab{}.
\newblock \showarticletitle{High-Resolution Daytime Translation Without Domain Labels}. In \bibinfo{booktitle}{\emph{The IEEE Conference on Computer Vision and Pattern Recognition (CVPR)}}.
\newblock


\bibitem[Chen et~al\mbox{.}(2021)]%
        {mvsnerf}
\bibfield{author}{\bibinfo{person}{Anpei Chen}, \bibinfo{person}{Zexiang Xu}, \bibinfo{person}{Fuqiang Zhao}, \bibinfo{person}{Xiaoshuai Zhang}, \bibinfo{person}{Fanbo Xiang}, \bibinfo{person}{Jingyi Yu}, {and} \bibinfo{person}{Hao Su}.} \bibinfo{year}{2021}\natexlab{}.
\newblock \showarticletitle{Mvsnerf: Fast generalizable radiance field reconstruction from multi-view stereo}. In \bibinfo{booktitle}{\emph{Proceedings of the IEEE/CVF International Conference on Computer Vision}}. \bibinfo{pages}{14124--14133}.
\newblock


\bibitem[Chen et~al\mbox{.}(2022b)]%
        {hanerf}
\bibfield{author}{\bibinfo{person}{Xingyu Chen}, \bibinfo{person}{Qi Zhang}, \bibinfo{person}{Xiaoyu Li}, \bibinfo{person}{Yue Chen}, \bibinfo{person}{Ying Feng}, \bibinfo{person}{Xuan Wang}, {and} \bibinfo{person}{Jue Wang}.} \bibinfo{year}{2022}\natexlab{b}.
\newblock \showarticletitle{Hallucinated neural radiance fields in the wild}. In \bibinfo{booktitle}{\emph{CVPR}}. \bibinfo{pages}{12943--12952}.
\newblock


\bibitem[Chen et~al\mbox{.}(2022a)]%
        {Time-of-Day}
\bibfield{author}{\bibinfo{person}{Yingshu Chen}, \bibinfo{person}{Tuan-Anh Vu}, \bibinfo{person}{Ka-Chun Shum}, \bibinfo{person}{Sai-Kit Yeung}, {and} \bibinfo{person}{Binh-Son Hua}.} \bibinfo{year}{2022}\natexlab{a}.
\newblock \showarticletitle{Time-of-Day Neural Style Transfer for Architectural Photographs}. In \bibinfo{booktitle}{\emph{2022 IEEE International Conference on Computational Photography (ICCP)}}. \bibinfo{pages}{1--12}.
\newblock
\urldef\tempurl%
\url{https://doi.org/10.1109/ICCP54855.2022.9887763}
\showDOI{\tempurl}


\bibitem[Cheng et~al\mbox{.}(2020)]%
        {Time_Flies}
\bibfield{author}{\bibinfo{person}{Chia-Chi Cheng}, \bibinfo{person}{Hung-Yu Chen}, {and} \bibinfo{person}{Wei-Chen Chiu}.} \bibinfo{year}{2020}\natexlab{}.
\newblock \showarticletitle{Time Flies: Animating a Still Image With Time-Lapse Video As Reference}. In \bibinfo{booktitle}{\emph{IEEE/CVF Conference on Computer Vision and Pattern Recognition (CVPR)}}.
\newblock


\bibitem[Chiang et~al\mbox{.}(2022)]%
        {style3D}
\bibfield{author}{\bibinfo{person}{Pei-Ze Chiang}, \bibinfo{person}{Meng-Shiun Tsai}, \bibinfo{person}{Hung-Yu Tseng}, \bibinfo{person}{Wei-Sheng Lai}, {and} \bibinfo{person}{Wei-Chen Chiu}.} \bibinfo{year}{2022}\natexlab{}.
\newblock \showarticletitle{Stylizing 3D Scene via Implicit Representation and HyperNetwork}. In \bibinfo{booktitle}{\emph{Proceedings of the IEEE/CVF Winter Conference on Applications of Computer Vision (WACV)}}.
\newblock


\bibitem[Deng and Tartaglione(2023)]%
        {evgnerf}
\bibfield{author}{\bibinfo{person}{Chenxi~Lola Deng} {and} \bibinfo{person}{Enzo Tartaglione}.} \bibinfo{year}{2023}\natexlab{}.
\newblock \showarticletitle{Compressing Explicit Voxel Grid Representations: Fast NeRFs Become Also Small}. In \bibinfo{booktitle}{\emph{Proceedings of the IEEE/CVF Winter Conference on Applications of Computer Vision (WACV)}}. \bibinfo{pages}{1236--1245}.
\newblock


\bibitem[Deng et~al\mbox{.}(2022)]%
        {dsnerf}
\bibfield{author}{\bibinfo{person}{Kangle Deng}, \bibinfo{person}{Andrew Liu}, \bibinfo{person}{Jun-Yan Zhu}, {and} \bibinfo{person}{Deva Ramanan}.} \bibinfo{year}{2022}\natexlab{}.
\newblock \showarticletitle{Depth-supervised {NeRF}: Fewer Views and Faster Training for Free}. In \bibinfo{booktitle}{\emph{Proceedings of the IEEE/CVF Conference on Computer Vision and Pattern Recognition (CVPR)}}.
\newblock


\bibitem[Diaz-Ruiz et~al\mbox{.}(2022)]%
        {ithaca}
\bibfield{author}{\bibinfo{person}{Carlos~A. Diaz-Ruiz}, \bibinfo{person}{Youya Xia}, \bibinfo{person}{Yurong You}, \bibinfo{person}{Jose Nino}, \bibinfo{person}{Junan Chen}, \bibinfo{person}{Josephine Monica}, \bibinfo{person}{Xiangyu Chen}, \bibinfo{person}{Katie Luo}, \bibinfo{person}{Yan Wang}, \bibinfo{person}{Marc Emond}, \bibinfo{person}{Wei-Lun Chao}, \bibinfo{person}{Bharath Hariharan}, \bibinfo{person}{Kilian~Q. Weinberger}, {and} \bibinfo{person}{Mark Campbell}.} \bibinfo{year}{2022}\natexlab{}.
\newblock \showarticletitle{Ithaca365: Dataset and Driving Perception Under Repeated and Challenging Weather Conditions}. In \bibinfo{booktitle}{\emph{Proceedings of the IEEE/CVF Conference on Computer Vision and Pattern Recognition (CVPR)}}. \bibinfo{pages}{21383--21392}.
\newblock


\bibitem[Gong et~al\mbox{.}(2019)]%
        {DLOW}
\bibfield{author}{\bibinfo{person}{Rui Gong}, \bibinfo{person}{Wen Li}, \bibinfo{person}{Yuhua Chen}, {and} \bibinfo{person}{Luc Van~Gool}.} \bibinfo{year}{2019}\natexlab{}.
\newblock \showarticletitle{DLOW: Domain Flow for Adaptation and Generalization}. In \bibinfo{booktitle}{\emph{2019 IEEE/CVF Conference on Computer Vision and Pattern Recognition (CVPR)}}. \bibinfo{pages}{2472--2481}.
\newblock
\urldef\tempurl%
\url{https://doi.org/10.1109/CVPR.2019.00258}
\showDOI{\tempurl}


\bibitem[Gu et~al\mbox{.}(2020)]%
        {cascade}
\bibfield{author}{\bibinfo{person}{Xiaodong Gu}, \bibinfo{person}{Zhiwen Fan}, \bibinfo{person}{Siyu Zhu}, \bibinfo{person}{Zuozhuo Dai}, \bibinfo{person}{Feitong Tan}, {and} \bibinfo{person}{Ping Tan}.} \bibinfo{year}{2020}\natexlab{}.
\newblock \showarticletitle{Cascade cost volume for high-resolution multi-view stereo and stereo matching}. In \bibinfo{booktitle}{\emph{Proceedings of the IEEE/CVF Conference on Computer Vision and Pattern Recognition}}. \bibinfo{pages}{2495--2504}.
\newblock


\bibitem[Guangcong et~al\mbox{.}(2023)]%
        {sparsenerf}
\bibfield{author}{\bibinfo{person}{Guangcong}, \bibinfo{person}{Zhaoxi Chen}, \bibinfo{person}{Chen~Change Loy}, {and} \bibinfo{person}{Ziwei Liu}.} \bibinfo{year}{2023}\natexlab{}.
\newblock \showarticletitle{SparseNeRF: Distilling Depth Ranking for Few-shot Novel View Synthesis}.
\newblock \bibinfo{journal}{\emph{Technical Report}} (\bibinfo{year}{2023}).
\newblock


\bibitem[Guo et~al\mbox{.}(2022)]%
        {DyNeRF}
\bibfield{author}{\bibinfo{person}{Haoyu Guo}, \bibinfo{person}{Sida Peng}, \bibinfo{person}{Haotong Lin}, \bibinfo{person}{Qianqian Wang}, \bibinfo{person}{Guofeng Zhang}, \bibinfo{person}{Hujun Bao}, {and} \bibinfo{person}{Xiaowei Zhou}.} \bibinfo{year}{2022}\natexlab{}.
\newblock \showarticletitle{Neural 3D Scene Reconstruction With the Manhattan-World Assumption}. In \bibinfo{booktitle}{\emph{CVPR}}. \bibinfo{pages}{5511--5520}.
\newblock


\bibitem[Hedman et~al\mbox{.}(2021)]%
        {SNeRG}
\bibfield{author}{\bibinfo{person}{Peter Hedman}, \bibinfo{person}{Pratul~P. Srinivasan}, \bibinfo{person}{Ben Mildenhall}, \bibinfo{person}{Jonathan~T. Barron}, {and} \bibinfo{person}{Paul Debevec}.} \bibinfo{year}{2021}\natexlab{}.
\newblock \showarticletitle{Baking Neural Radiance Fields for Real-Time View Synthesis}.
\newblock \bibinfo{journal}{\emph{ICCV}} (\bibinfo{year}{2021}).
\newblock


\bibitem[Huang et~al\mbox{.}(2021)]%
        {learnStyle}
\bibfield{author}{\bibinfo{person}{Hsin-Ping Huang}, \bibinfo{person}{Hung-Yu Tseng}, \bibinfo{person}{Saurabh Saini}, \bibinfo{person}{Maneesh Singh}, {and} \bibinfo{person}{Ming-Hsuan Yang}.} \bibinfo{year}{2021}\natexlab{}.
\newblock \showarticletitle{Learning to Stylize Novel Views}. In \bibinfo{booktitle}{\emph{ICCV}}.
\newblock


\bibitem[Huang et~al\mbox{.}(2022)]%
        {Huang22StylizedNeRF}
\bibfield{author}{\bibinfo{person}{Yi-Hua Huang}, \bibinfo{person}{Yue He}, \bibinfo{person}{Yu-Jie Yuan}, \bibinfo{person}{Yu-Kun Lai}, {and} \bibinfo{person}{Lin Gao}.} \bibinfo{year}{2022}\natexlab{}.
\newblock \showarticletitle{StylizedNeRF: Consistent 3D Scene Stylization as Stylized NeRF via 2D-3D Mutual Learning}. In \bibinfo{booktitle}{\emph{Computer Vision and Pattern Recognition (CVPR)}}.
\newblock


\bibitem[Jain et~al\mbox{.}(2021)]%
        {dietNeRF}
\bibfield{author}{\bibinfo{person}{Ajay Jain}, \bibinfo{person}{Matthew Tancik}, {and} \bibinfo{person}{Pieter Abbeel}.} \bibinfo{year}{2021}\natexlab{}.
\newblock \showarticletitle{Putting NeRF on a Diet: Semantically Consistent Few-Shot View Synthesis}. In \bibinfo{booktitle}{\emph{Proceedings of the IEEE/CVF International Conference on Computer Vision (ICCV)}}. \bibinfo{pages}{5885--5894}.
\newblock


\bibitem[Jiang et~al\mbox{.}(2020)]%
        {tsit}
\bibfield{author}{\bibinfo{person}{Liming Jiang}, \bibinfo{person}{Changxu Zhang}, \bibinfo{person}{Mingyang Huang}, \bibinfo{person}{Chunxiao Liu}, \bibinfo{person}{Jianping Shi}, {and} \bibinfo{person}{Chen~Change Loy}.} \bibinfo{year}{2020}\natexlab{}.
\newblock \showarticletitle{{TSIT}: A Simple and Versatile Framework for Image-to-Image Translation}. In \bibinfo{booktitle}{\emph{ECCV}}.
\newblock


\bibitem[Johari et~al\mbox{.}(2022)]%
        {GeoNeRF}
\bibfield{author}{\bibinfo{person}{M. Johari}, \bibinfo{person}{Y. Lepoittevin}, {and} \bibinfo{person}{F. Fleuret}.} \bibinfo{year}{2022}\natexlab{}.
\newblock \showarticletitle{GeoNeRF: Generalizing NeRF with Geometry Priors}. In \bibinfo{booktitle}{\emph{Proceedings of the IEEE International Conference on Computer Vision and Pattern Recognition (CVPR)}}.
\newblock


\bibitem[Kim et~al\mbox{.}(2022)]%
        {infonerf}
\bibfield{author}{\bibinfo{person}{Mijeong Kim}, \bibinfo{person}{Seonguk Seo}, {and} \bibinfo{person}{Bohyung Han}.} \bibinfo{year}{2022}\natexlab{}.
\newblock \showarticletitle{InfoNeRF: Ray Entropy Minimization for Few-Shot Neural Volume Rendering}. In \bibinfo{booktitle}{\emph{CVPR}}.
\newblock


\bibitem[Kingma and Ba(2015)]%
        {adam}
\bibfield{author}{\bibinfo{person}{Diederik Kingma} {and} \bibinfo{person}{Jimmy Ba}.} \bibinfo{year}{2015}\natexlab{}.
\newblock \showarticletitle{Adam: A Method for Stochastic Optimization}. In \bibinfo{booktitle}{\emph{International Conference on Learning Representations (ICLR)}}. \bibinfo{address}{San Diega, CA, USA}.
\newblock


\bibitem[Knapitsch et~al\mbox{.}(2017)]%
        {t&t}
\bibfield{author}{\bibinfo{person}{Arno Knapitsch}, \bibinfo{person}{Jaesik Park}, \bibinfo{person}{Qian-Yi Zhou}, {and} \bibinfo{person}{Vladlen Koltun}.} \bibinfo{year}{2017}\natexlab{}.
\newblock \showarticletitle{Tanks and Temples: Benchmarking Large-Scale Scene Reconstruction}.
\newblock \bibinfo{journal}{\emph{ACM Transactions on Graphics}} \bibinfo{volume}{36}, \bibinfo{number}{4} (\bibinfo{year}{2017}).
\newblock


\bibitem[Kuang et~al\mbox{.}(2022)]%
        {NeROIC}
\bibfield{author}{\bibinfo{person}{Zhengfei Kuang}, \bibinfo{person}{Kyle Olszewski}, \bibinfo{person}{Menglei Chai}, \bibinfo{person}{Zeng Huang}, \bibinfo{person}{Panos Achlioptas}, {and} \bibinfo{person}{Sergey Tulyakov}.} \bibinfo{year}{2022}\natexlab{}.
\newblock \showarticletitle{NeROIC: Neural Rendering of Objects from Online Image Collections}.
\newblock \bibinfo{journal}{\emph{ACM Trans. Graph.}} \bibinfo{volume}{41}, \bibinfo{number}{4}, Article \bibinfo{articleno}{56} (\bibinfo{date}{jul} \bibinfo{year}{2022}), \bibinfo{numpages}{12}~pages.
\newblock
\showISSN{0730-0301}
\urldef\tempurl%
\url{https://doi.org/10.1145/3528223.3530177}
\showDOI{\tempurl}


\bibitem[Kurz et~al\mbox{.}(2022)]%
        {adanerf2022}
\bibfield{author}{\bibinfo{person}{Andreas Kurz}, \bibinfo{person}{Thomas Neff}, \bibinfo{person}{Zhaoyang Lv}, \bibinfo{person}{Michael Zollh\"{o}fer}, {and} \bibinfo{person}{Markus Steinberger}.} \bibinfo{year}{2022}\natexlab{}.
\newblock \showarticletitle{AdaNeRF: Adaptive Sampling for Real-time Rendering of Neural Radiance Fields}.
\newblock  (\bibinfo{year}{2022}).
\newblock


\bibitem[Lee et~al\mbox{.}(2018)]%
        {DRIT}
\bibfield{author}{\bibinfo{person}{Hsin-Ying Lee}, \bibinfo{person}{Hung-Yu Tseng}, \bibinfo{person}{Jia-Bin Huang}, \bibinfo{person}{Maneesh~Kumar Singh}, {and} \bibinfo{person}{Ming-Hsuan Yang}.} \bibinfo{year}{2018}\natexlab{}.
\newblock \showarticletitle{Diverse Image-to-Image Translation via Disentangled Representations}. In \bibinfo{booktitle}{\emph{European Conference on Computer Vision}}.
\newblock


\bibitem[Lee et~al\mbox{.}(2020)]%
        {DRIT_plus}
\bibfield{author}{\bibinfo{person}{Hsin-Ying Lee}, \bibinfo{person}{Hung-Yu Tseng}, \bibinfo{person}{Qi Mao}, \bibinfo{person}{Jia-Bin Huang}, \bibinfo{person}{Yu-Ding Lu}, \bibinfo{person}{Maneesh~Kumar Singh}, {and} \bibinfo{person}{Ming-Hsuan Yang}.} \bibinfo{year}{2020}\natexlab{}.
\newblock \showarticletitle{DRIT++: Diverse Image-to-Image Translation via Disentangled Representations}.
\newblock \bibinfo{journal}{\emph{International Journal of Computer Vision}} (\bibinfo{year}{2020}), \bibinfo{pages}{1--16}.
\newblock


\bibitem[Lin et~al\mbox{.}(2023)]%
        {visionnerf}
\bibfield{author}{\bibinfo{person}{Kai-En Lin}, \bibinfo{person}{Lin Yen-Chen}, \bibinfo{person}{Wei-Sheng Lai}, \bibinfo{person}{Tsung-Yi Lin}, \bibinfo{person}{Yi-Chang Shih}, {and} \bibinfo{person}{Ravi Ramamoorthi}.} \bibinfo{year}{2023}\natexlab{}.
\newblock \showarticletitle{Vision Transformer for NeRF-Based View Synthesis from a Single Input Image}. In \bibinfo{booktitle}{\emph{WACV}}.
\newblock


\bibitem[Lindell et~al\mbox{.}(2021)]%
        {autoint2021}
\bibfield{author}{\bibinfo{person}{David~B. Lindell}, \bibinfo{person}{Julien N.~P. Martel}, {and} \bibinfo{person}{Gordon Wetzstein}.} \bibinfo{year}{2021}\natexlab{}.
\newblock \showarticletitle{AutoInt: Automatic Integration for Fast Neural Volume Rendering}. In \bibinfo{booktitle}{\emph{Proc. CVPR}}.
\newblock


\bibitem[Liu et~al\mbox{.}(2020)]%
        {NSVF}
\bibfield{author}{\bibinfo{person}{Lingjie Liu}, \bibinfo{person}{Jiatao Gu}, \bibinfo{person}{Kyaw~Zaw Lin}, \bibinfo{person}{Tat-Seng Chua}, {and} \bibinfo{person}{Christian Theobalt}.} \bibinfo{year}{2020}\natexlab{}.
\newblock \showarticletitle{Neural Sparse Voxel Fields}.
\newblock \bibinfo{journal}{\emph{NeurIPS}} (\bibinfo{year}{2020}).
\newblock


\bibitem[Liu et~al\mbox{.}(2022)]%
        {neuray}
\bibfield{author}{\bibinfo{person}{Yuan Liu}, \bibinfo{person}{Sida Peng}, \bibinfo{person}{Lingjie Liu}, \bibinfo{person}{Qianqian Wang}, \bibinfo{person}{Peng Wang}, \bibinfo{person}{Christian Theobalt}, \bibinfo{person}{Xiaowei Zhou}, {and} \bibinfo{person}{Wenping Wang}.} \bibinfo{year}{2022}\natexlab{}.
\newblock \showarticletitle{Neural Rays for Occlusion-aware Image-based Rendering}. In \bibinfo{booktitle}{\emph{CVPR}}.
\newblock


\bibitem[Lombardi et~al\mbox{.}(2019)]%
        {NeuralVolumes}
\bibfield{author}{\bibinfo{person}{Stephen Lombardi}, \bibinfo{person}{Tomas Simon}, \bibinfo{person}{Jason Saragih}, \bibinfo{person}{Gabriel Schwartz}, \bibinfo{person}{Andreas Lehrmann}, {and} \bibinfo{person}{Yaser Sheikh}.} \bibinfo{year}{2019}\natexlab{}.
\newblock \showarticletitle{Neural Volumes: Learning Dynamic Renderable Volumes from Images}.
\newblock \bibinfo{journal}{\emph{ACM Trans. Graph.}} \bibinfo{volume}{38}, \bibinfo{number}{4}, Article \bibinfo{articleno}{65} (\bibinfo{date}{July} \bibinfo{year}{2019}), \bibinfo{numpages}{14}~pages.
\newblock


\bibitem[Loshchilov and Hutter(2017)]%
        {sgd}
\bibfield{author}{\bibinfo{person}{Ilya Loshchilov} {and} \bibinfo{person}{Frank Hutter}.} \bibinfo{year}{2017}\natexlab{}.
\newblock \showarticletitle{{SGDR}: Stochastic Gradient Descent with Warm Restarts}. In \bibinfo{booktitle}{\emph{International Conference on Learning Representations}}.
\newblock
\urldef\tempurl%
\url{https://openreview.net/forum?id=Skq89Scxx}
\showURL{%
\tempurl}


\bibitem[Martin-Brualla et~al\mbox{.}(2021)]%
        {nerfw}
\bibfield{author}{\bibinfo{person}{Ricardo Martin-Brualla}, \bibinfo{person}{Noha Radwan}, \bibinfo{person}{Mehdi S.~M. Sajjadi}, \bibinfo{person}{Jonathan~T. Barron}, \bibinfo{person}{Alexey Dosovitskiy}, {and} \bibinfo{person}{Daniel Duckworth}.} \bibinfo{year}{2021}\natexlab{}.
\newblock \showarticletitle{{NeRF in the Wild: Neural Radiance Fields for Unconstrained Photo Collections}}. In \bibinfo{booktitle}{\emph{CVPR}}.
\newblock


\bibitem[Mildenhall et~al\mbox{.}(2019)]%
        {llff}
\bibfield{author}{\bibinfo{person}{Ben Mildenhall}, \bibinfo{person}{Pratul~P. Srinivasan}, \bibinfo{person}{Rodrigo Ortiz-Cayon}, \bibinfo{person}{Nima~Khademi Kalantari}, \bibinfo{person}{Ravi Ramamoorthi}, \bibinfo{person}{Ren Ng}, {and} \bibinfo{person}{Abhishek Kar}.} \bibinfo{year}{2019}\natexlab{}.
\newblock \showarticletitle{Local Light Field Fusion: Practical View Synthesis with Prescriptive Sampling Guidelines}.
\newblock \bibinfo{journal}{\emph{ACM Trans. Graph.}} \bibinfo{volume}{38}, \bibinfo{number}{4}, Article \bibinfo{articleno}{29} (\bibinfo{date}{jul} \bibinfo{year}{2019}), \bibinfo{numpages}{14}~pages.
\newblock
\showISSN{0730-0301}
\urldef\tempurl%
\url{https://doi.org/10.1145/3306346.3322980}
\showDOI{\tempurl}


\bibitem[Mildenhall et~al\mbox{.}(2020)]%
        {nerf}
\bibfield{author}{\bibinfo{person}{Ben Mildenhall}, \bibinfo{person}{Pratul~P. Srinivasan}, \bibinfo{person}{Matthew Tancik}, \bibinfo{person}{Jonathan~T. Barron}, \bibinfo{person}{Ravi Ramamoorthi}, {and} \bibinfo{person}{Ren Ng}.} \bibinfo{year}{2020}\natexlab{}.
\newblock \showarticletitle{NeRF: Representing Scenes as Neural Radiance Fields for View Synthesis}. In \bibinfo{booktitle}{\emph{ECCV}}.
\newblock


\bibitem[M\"uller et~al\mbox{.}(2022)]%
        {instantngp}
\bibfield{author}{\bibinfo{person}{Thomas M\"uller}, \bibinfo{person}{Alex Evans}, \bibinfo{person}{Christoph Schied}, {and} \bibinfo{person}{Alexander Keller}.} \bibinfo{year}{2022}\natexlab{}.
\newblock \showarticletitle{Instant Neural Graphics Primitives with a Multiresolution Hash Encoding}.
\newblock \bibinfo{journal}{\emph{ACM Trans. Graph.}} \bibinfo{volume}{41}, \bibinfo{number}{4}, Article \bibinfo{articleno}{102} (\bibinfo{date}{July} \bibinfo{year}{2022}), \bibinfo{numpages}{15}~pages.
\newblock
\urldef\tempurl%
\url{https://doi.org/10.1145/3528223.3530127}
\showDOI{\tempurl}


\bibitem[Nam et~al\mbox{.}(2019)]%
        {End-To-End_Time-Lapse}
\bibfield{author}{\bibinfo{person}{Seonghyeon Nam}, \bibinfo{person}{Chongyang Ma}, \bibinfo{person}{Menglei Chai}, \bibinfo{person}{William Brendel}, \bibinfo{person}{Ning Xu}, {and} \bibinfo{person}{Seon~Joo Kim}.} \bibinfo{year}{2019}\natexlab{}.
\newblock \showarticletitle{End-To-End Time-Lapse Video Synthesis From a Single Outdoor Image}. In \bibinfo{booktitle}{\emph{Proceedings of the IEEE/CVF Conference on Computer Vision and Pattern Recognition (CVPR)}}.
\newblock


\bibitem[Neff et~al\mbox{.}(2021)]%
        {donerf}
\bibfield{author}{\bibinfo{person}{T. Neff}, \bibinfo{person}{P. Stadlbauer}, \bibinfo{person}{M. Parger}, \bibinfo{person}{A. Kurz}, \bibinfo{person}{J.~H. Mueller}, \bibinfo{person}{C.~R.~A. Chaitanya}, \bibinfo{person}{A. Kaplanyan}, {and} \bibinfo{person}{M. Steinberger}.} \bibinfo{year}{2021}\natexlab{}.
\newblock \showarticletitle{DONeRF: Towards Real-Time Rendering of Compact Neural Radiance Fields using Depth Oracle Networks}.
\newblock \bibinfo{journal}{\emph{Computer Graphics Forum}} \bibinfo{volume}{40}, \bibinfo{number}{4} (\bibinfo{year}{2021}), \bibinfo{pages}{45--59}.
\newblock
\urldef\tempurl%
\url{https://doi.org/10.1111/cgf.14340}
\showDOI{\tempurl}
\showeprint{https://onlinelibrary.wiley.com/doi/pdf/10.1111/cgf.14340}


\bibitem[Nguyen-Ha et~al\mbox{.}(2021)]%
        {rgbdNet}
\bibfield{author}{\bibinfo{person}{Phong Nguyen-Ha}, \bibinfo{person}{Animesh Karnewar}, \bibinfo{person}{Lam Huynh}, \bibinfo{person}{Esa Rahtu}, {and} \bibinfo{person}{Janne Heikkila}.} \bibinfo{year}{2021}\natexlab{}.
\newblock \showarticletitle{RGBD-Net: Predicting color and depth images for novel views synthesis}. In \bibinfo{booktitle}{\emph{Proceedings of the International Conference on 3D Vision}}.
\newblock


\bibitem[Niemeyer et~al\mbox{.}(2022)]%
        {Regnerf}
\bibfield{author}{\bibinfo{person}{Michael Niemeyer}, \bibinfo{person}{Jonathan~T. Barron}, \bibinfo{person}{Ben Mildenhall}, \bibinfo{person}{Mehdi S.~M. Sajjadi}, \bibinfo{person}{Andreas Geiger}, {and} \bibinfo{person}{Noha Radwan}.} \bibinfo{year}{2022}\natexlab{}.
\newblock \showarticletitle{RegNeRF: Regularizing Neural Radiance Fields for View Synthesis from Sparse Inputs}. In \bibinfo{booktitle}{\emph{Proc. IEEE Conf. on Computer Vision and Pattern Recognition (CVPR)}}.
\newblock


\bibitem[Park et~al\mbox{.}(2021)]%
        {hypernerf}
\bibfield{author}{\bibinfo{person}{Keunhong Park}, \bibinfo{person}{Utkarsh Sinha}, \bibinfo{person}{Peter Hedman}, \bibinfo{person}{Jonathan~T. Barron}, \bibinfo{person}{Sofien Bouaziz}, \bibinfo{person}{Dan~B Goldman}, \bibinfo{person}{Ricardo Martin-Brualla}, {and} \bibinfo{person}{Steven~M. Seitz}.} \bibinfo{year}{2021}\natexlab{}.
\newblock \showarticletitle{HyperNeRF: A Higher-Dimensional Representation for Topologically Varying Neural Radiance Fields}.
\newblock \bibinfo{journal}{\emph{ACM Trans. Graph.}} \bibinfo{volume}{40}, \bibinfo{number}{6}, Article \bibinfo{articleno}{238} (\bibinfo{date}{dec} \bibinfo{year}{2021}).
\newblock


\bibitem[Piala and Clark(2021)]%
        {TermiNeRF}
\bibfield{author}{\bibinfo{person}{Martin Piala} {and} \bibinfo{person}{Ronald Clark}.} \bibinfo{year}{2021}\natexlab{}.
\newblock \showarticletitle{TermiNeRF: Ray Termination Prediction for Efficient Neural Rendering}. In \bibinfo{booktitle}{\emph{2021 International Conference on 3D Vision (3DV)}}. \bibinfo{pages}{1106--1114}.
\newblock
\urldef\tempurl%
\url{https://doi.org/10.1109/3DV53792.2021.00118}
\showDOI{\tempurl}


\bibitem[Pizzati et~al\mbox{.}(2021)]%
        {comogan}
\bibfield{author}{\bibinfo{person}{Fabio Pizzati}, \bibinfo{person}{Pietro Cerri}, {and} \bibinfo{person}{Raoul de Charette}.} \bibinfo{year}{2021}\natexlab{}.
\newblock \showarticletitle{{CoMoGAN}: continuous model-guided image-to-image translation}. In \bibinfo{booktitle}{\emph{CVPR}}.
\newblock


\bibitem[Pumarola et~al\mbox{.}(2020)]%
        {d-nerf}
\bibfield{author}{\bibinfo{person}{Albert Pumarola}, \bibinfo{person}{Enric Corona}, \bibinfo{person}{Gerard Pons-Moll}, {and} \bibinfo{person}{Francesc Moreno-Noguer}.} \bibinfo{year}{2020}\natexlab{}.
\newblock \showarticletitle{D-NeRF: Neural Radiance Fields for Dynamic Scenes}. In \bibinfo{booktitle}{\emph{CVPR}}.
\newblock


\bibitem[Rafique et~al\mbox{.}(2021)]%
        {UnifyingG}
\bibfield{author}{\bibinfo{person}{Muhammad~Usman Rafique}, \bibinfo{person}{Yu Zhang}, \bibinfo{person}{Benjamin Brodie}, {and} \bibinfo{person}{Nathan Jacobs}.} \bibinfo{year}{2021}\natexlab{}.
\newblock \showarticletitle{Unifying Guided and Unguided Outdoor Image Synthesis}. In \bibinfo{booktitle}{\emph{2021 IEEE/CVF Conference on Computer Vision and Pattern Recognition Workshops (CVPRW)}}. \bibinfo{pages}{776--785}.
\newblock
\urldef\tempurl%
\url{https://doi.org/10.1109/CVPRW53098.2021.00087}
\showDOI{\tempurl}


\bibitem[Roessle et~al\mbox{.}(2022)]%
        {depthpriorsnerf}
\bibfield{author}{\bibinfo{person}{Barbara Roessle}, \bibinfo{person}{Jonathan~T. Barron}, \bibinfo{person}{Ben Mildenhall}, \bibinfo{person}{Pratul~P. Srinivasan}, {and} \bibinfo{person}{Matthias Nie{\ss}ner}.} \bibinfo{year}{2022}\natexlab{}.
\newblock \showarticletitle{Dense Depth Priors for Neural Radiance Fields from Sparse Input Views}. In \bibinfo{booktitle}{\emph{Proceedings of the IEEE/CVF Conference on Computer Vision and Pattern Recognition (CVPR)}}.
\newblock


\bibitem[Romero et~al\mbox{.}(2019)]%
        {smit}
\bibfield{author}{\bibinfo{person}{Andr{\'e}s Romero}, \bibinfo{person}{Pablo Arbel{\'a}ez}, \bibinfo{person}{Luc Van~Gool}, {and} \bibinfo{person}{Radu Timofte}.} \bibinfo{year}{2019}\natexlab{}.
\newblock \showarticletitle{SMIT: Stochastic Multi-Label Image-to-Image Translation}.
\newblock \bibinfo{journal}{\emph{ICCV Workshops}} (\bibinfo{year}{2019}).
\newblock


\bibitem[{Sara Fridovich-Keil and Giacomo Meanti} et~al\mbox{.}(2023)]%
        {kplanes_2023}
\bibfield{author}{\bibinfo{person}{{Sara Fridovich-Keil and Giacomo Meanti}}, \bibinfo{person}{Frederik~Rahbæk Warburg}, \bibinfo{person}{Benjamin Recht}, {and} \bibinfo{person}{Angjoo Kanazawa}.} \bibinfo{year}{2023}\natexlab{}.
\newblock \showarticletitle{K-Planes: Explicit Radiance Fields in Space, Time, and Appearance}. In \bibinfo{booktitle}{\emph{CVPR}}.
\newblock


\bibitem[Seo et~al\mbox{.}(2023)]%
        {mixnerf}
\bibfield{author}{\bibinfo{person}{Seunghyeon Seo}, \bibinfo{person}{Donghoon Han}, \bibinfo{person}{Yeonjin Chang}, {and} \bibinfo{person}{Nojun Kwak}.} \bibinfo{year}{2023}\natexlab{}.
\newblock \showarticletitle{MixNeRF: Modeling a Ray With Mixture Density for Novel View Synthesis From Sparse Inputs}. In \bibinfo{booktitle}{\emph{Proceedings of the IEEE/CVF Conference on Computer Vision and Pattern Recognition (CVPR)}}. \bibinfo{pages}{20659--20668}.
\newblock


\bibitem[Sun et~al\mbox{.}(2022)]%
        {DVGO}
\bibfield{author}{\bibinfo{person}{Cheng Sun}, \bibinfo{person}{Min Sun}, {and} \bibinfo{person}{Hwann{-}Tzong Chen}.} \bibinfo{year}{2022}\natexlab{}.
\newblock \showarticletitle{Direct Voxel Grid Optimization: Super-fast Convergence for Radiance Fields Reconstruction}. In \bibinfo{booktitle}{\emph{CVPR}}.
\newblock


\bibitem[Sun et~al\mbox{.}(2020)]%
        {waymo}
\bibfield{author}{\bibinfo{person}{Pei Sun}, \bibinfo{person}{Henrik Kretzschmar}, \bibinfo{person}{Xerxes Dotiwalla}, \bibinfo{person}{Aurelien Chouard}, \bibinfo{person}{Vijaysai Patnaik}, \bibinfo{person}{Paul Tsui}, \bibinfo{person}{James Guo}, \bibinfo{person}{Yin Zhou}, \bibinfo{person}{Yuning Chai}, \bibinfo{person}{Benjamin Caine}, \bibinfo{person}{Vijay Vasudevan}, \bibinfo{person}{Wei Han}, \bibinfo{person}{Jiquan Ngiam}, \bibinfo{person}{Hang Zhao}, \bibinfo{person}{Aleksei Timofeev}, \bibinfo{person}{Scott Ettinger}, \bibinfo{person}{Maxim Krivokon}, \bibinfo{person}{Amy Gao}, \bibinfo{person}{Aditya Joshi}, \bibinfo{person}{Yu Zhang}, \bibinfo{person}{Jonathon Shlens}, \bibinfo{person}{Zhifeng Chen}, {and} \bibinfo{person}{Dragomir Anguelov}.} \bibinfo{year}{2020}\natexlab{}.
\newblock \showarticletitle{Scalability in Perception for Autonomous Driving: Waymo Open Dataset}. In \bibinfo{booktitle}{\emph{Proceedings of the IEEE/CVF Conference on Computer Vision and Pattern Recognition (CVPR)}}.
\newblock


\bibitem[Tancik et~al\mbox{.}(2022)]%
        {Block-NeRF}
\bibfield{author}{\bibinfo{person}{Matthew Tancik}, \bibinfo{person}{Vincent Casser}, \bibinfo{person}{Xinchen Yan}, \bibinfo{person}{Sabeek Pradhan}, \bibinfo{person}{Ben Mildenhall}, \bibinfo{person}{Pratul~P. Srinivasan}, \bibinfo{person}{Jonathan~T. Barron}, {and} \bibinfo{person}{Henrik Kretzschmar}.} \bibinfo{year}{2022}\natexlab{}.
\newblock \showarticletitle{Block-NeRF: Scalable Large Scene Neural View Synthesis}. In \bibinfo{booktitle}{\emph{CVPR}}. \bibinfo{pages}{8248--8258}.
\newblock


\bibitem[Vaswani et~al\mbox{.}(2017)]%
        {transformer}
\bibfield{author}{\bibinfo{person}{Ashish Vaswani}, \bibinfo{person}{Noam Shazeer}, \bibinfo{person}{Niki Parmar}, \bibinfo{person}{Jakob Uszkoreit}, \bibinfo{person}{Llion Jones}, \bibinfo{person}{Aidan~N. Gomez}, \bibinfo{person}{\L{}ukasz Kaiser}, {and} \bibinfo{person}{Illia Polosukhin}.} \bibinfo{year}{2017}\natexlab{}.
\newblock \showarticletitle{Attention is All You Need}. In \bibinfo{booktitle}{\emph{Proceedings of the 31st International Conference on Neural Information Processing Systems}} (Long Beach, California, USA) \emph{(\bibinfo{series}{NIPS'17})}. \bibinfo{publisher}{Curran Associates Inc.}, \bibinfo{address}{Red Hook, NY, USA}, \bibinfo{pages}{6000–6010}.
\newblock
\showISBNx{9781510860964}


\bibitem[Wang et~al\mbox{.}(2021)]%
        {ibrnet}
\bibfield{author}{\bibinfo{person}{Qianqian Wang}, \bibinfo{person}{Zhicheng Wang}, \bibinfo{person}{Kyle Genova}, \bibinfo{person}{Pratul Srinivasan}, \bibinfo{person}{Howard Zhou}, \bibinfo{person}{Jonathan~T. Barron}, \bibinfo{person}{Ricardo Martin-Brualla}, \bibinfo{person}{Noah Snavely}, {and} \bibinfo{person}{Thomas Funkhouser}.} \bibinfo{year}{2021}\natexlab{}.
\newblock \showarticletitle{IBRNet: Learning Multi-View Image-Based Rendering}. In \bibinfo{booktitle}{\emph{CVPR}}.
\newblock


\bibitem[Wang et~al\mbox{.}(2019)]%
        {DNI}
\bibfield{author}{\bibinfo{person}{Xintao Wang}, \bibinfo{person}{Ke Yu}, \bibinfo{person}{Chao Dong}, \bibinfo{person}{Xiaoou Tang}, {and} \bibinfo{person}{Chen~Change Loy}.} \bibinfo{year}{2019}\natexlab{}.
\newblock \showarticletitle{Deep Network Interpolation for Continuous Imagery Effect Transition}. In \bibinfo{booktitle}{\emph{2019 IEEE/CVF Conference on Computer Vision and Pattern Recognition (CVPR)}}. \bibinfo{pages}{1692--1701}.
\newblock
\urldef\tempurl%
\url{https://doi.org/10.1109/CVPR.2019.00179}
\showDOI{\tempurl}


\bibitem[Wu et~al\mbox{.}(2022)]%
        {DIVeR}
\bibfield{author}{\bibinfo{person}{Liwen Wu}, \bibinfo{person}{Jae~Yong Lee}, \bibinfo{person}{Anand Bhattad}, \bibinfo{person}{Yu-Xiong Wang}, {and} \bibinfo{person}{David Forsyth}.} \bibinfo{year}{2022}\natexlab{}.
\newblock \showarticletitle{DIVeR: Real-Time and Accurate Neural Radiance Fields With Deterministic Integration for Volume Rendering}. In \bibinfo{booktitle}{\emph{CVPR}}.
\newblock


\bibitem[Xin et~al\mbox{.}(2023)]%
        {lirf}
\bibfield{author}{\bibinfo{person}{Huang Xin}, \bibinfo{person}{Zhang Qi}, \bibinfo{person}{Feng Ying}, \bibinfo{person}{Li Xiaoyu}, \bibinfo{person}{Wang Xuan}, {and} \bibinfo{person}{Wang Qing}.} \bibinfo{year}{2023}\natexlab{}.
\newblock \showarticletitle{Local Implicit Ray Function for Generalizable Radiance Field Representation}. In \bibinfo{booktitle}{\emph{Proceedings of the IEEE/CVF Conference on Computer Vision and Pattern Recognition}}.
\newblock


\bibitem[Xu et~al\mbox{.}(2022)]%
        {sinnerf}
\bibfield{author}{\bibinfo{person}{Dejia Xu}, \bibinfo{person}{Yifan Jiang}, \bibinfo{person}{Peihao Wang}, \bibinfo{person}{Zhiwen Fan}, \bibinfo{person}{Humphrey Shi}, {and} \bibinfo{person}{Zhangyang Wang}.} \bibinfo{year}{2022}\natexlab{}.
\newblock \showarticletitle{SinNeRF: Training Neural Radiance Fields on Complex Scenes from a Single Image}. In \bibinfo{booktitle}{\emph{Computer Vision -- ECCV 2022}}, \bibfield{editor}{\bibinfo{person}{Shai Avidan}, \bibinfo{person}{Gabriel Brostow}, \bibinfo{person}{Moustapha Ciss{\'e}}, \bibinfo{person}{Giovanni~Maria Farinella}, {and} \bibinfo{person}{Tal Hassner}} (Eds.). \bibinfo{publisher}{Springer Nature Switzerland}, \bibinfo{address}{Cham}, \bibinfo{pages}{736--753}.
\newblock
\showISBNx{978-3-031-20047-2}


\bibitem[Xue et~al\mbox{.}(2021)]%
        {Landscape_Animation}
\bibfield{author}{\bibinfo{person}{Hongwei Xue}, \bibinfo{person}{Bei Liu}, \bibinfo{person}{Huan Yang}, \bibinfo{person}{Jianlong Fu}, \bibinfo{person}{Houqiang Li}, {and} \bibinfo{person}{Jiebo Luo}.} \bibinfo{year}{2021}\natexlab{}.
\newblock \showarticletitle{Learning Fine-Grained Motion Embedding for Landscape Animation}. In \bibinfo{booktitle}{\emph{Proceedings of the 29th ACM International Conference on Multimedia}} (Virtual Event, China) \emph{(\bibinfo{series}{MM '21})}. \bibinfo{publisher}{Association for Computing Machinery}, \bibinfo{address}{New York, NY, USA}, \bibinfo{pages}{291–299}.
\newblock
\showISBNx{9781450386517}
\urldef\tempurl%
\url{https://doi.org/10.1145/3474085.3475421}
\showDOI{\tempurl}


\bibitem[Yang et~al\mbox{.}(2023a)]%
        {ContraNeRF}
\bibfield{author}{\bibinfo{person}{Hao Yang}, \bibinfo{person}{Lanqing Hong}, \bibinfo{person}{Aoxue Li}, \bibinfo{person}{Tianyang Hu}, \bibinfo{person}{Zhenguo Li}, \bibinfo{person}{Gim~Hee Lee}, {and} \bibinfo{person}{Liwei Wang}.} \bibinfo{year}{2023}\natexlab{a}.
\newblock \showarticletitle{ContraNeRF: Generalizable Neural Radiance Fields for Synthetic-to-Real Novel View Synthesis via Contrastive Learning}. In \bibinfo{booktitle}{\emph{Proceedings of the IEEE/CVF Conference on Computer Vision and Pattern Recognition (CVPR)}}. \bibinfo{pages}{16508--16517}.
\newblock


\bibitem[Yang et~al\mbox{.}(2023b)]%
        {freenerf}
\bibfield{author}{\bibinfo{person}{Jiawei Yang}, \bibinfo{person}{Marco Pavone}, {and} \bibinfo{person}{Yue Wang}.} \bibinfo{year}{2023}\natexlab{b}.
\newblock \showarticletitle{FreeNeRF: Improving Few-shot Neural Rendering with Free Frequency Regularization}.
\newblock  (\bibinfo{year}{2023}).
\newblock


\bibitem[Yao et~al\mbox{.}(2018)]%
        {mvsnet}
\bibfield{author}{\bibinfo{person}{Yao Yao}, \bibinfo{person}{Zixin Luo}, \bibinfo{person}{Shiwei Li}, \bibinfo{person}{Tian Fang}, {and} \bibinfo{person}{Long Quan}.} \bibinfo{year}{2018}\natexlab{}.
\newblock \showarticletitle{MVSNet: Depth Inference for Unstructured Multi-view Stereo}.
\newblock \bibinfo{journal}{\emph{European Conference on Computer Vision (ECCV)}} (\bibinfo{year}{2018}).
\newblock


\bibitem[Yu et~al\mbox{.}(2021)]%
        {pixelnerf}
\bibfield{author}{\bibinfo{person}{Alex Yu}, \bibinfo{person}{Vickie Ye}, \bibinfo{person}{Matthew Tancik}, {and} \bibinfo{person}{Angjoo Kanazawa}.} \bibinfo{year}{2021}\natexlab{}.
\newblock \showarticletitle{{pixelNeRF}: Neural Radiance Fields from One or Few Images}. In \bibinfo{booktitle}{\emph{CVPR}}.
\newblock


\bibitem[Zhang et~al\mbox{.}(2020)]%
        {nerf++}
\bibfield{author}{\bibinfo{person}{Kai Zhang}, \bibinfo{person}{Gernot Riegler}, \bibinfo{person}{Noah Snavely}, {and} \bibinfo{person}{Vladlen Koltun}.} \bibinfo{year}{2020}\natexlab{}.
\newblock \showarticletitle{NeRF++: Analyzing and Improving Neural Radiance Fields}.
\newblock \bibinfo{journal}{\emph{arXiv:2010.07492}} (\bibinfo{year}{2020}).
\newblock


\bibitem[Zhang et~al\mbox{.}(2018)]%
        {lpips}
\bibfield{author}{\bibinfo{person}{Richard Zhang}, \bibinfo{person}{Phillip Isola}, \bibinfo{person}{Alexei~A Efros}, \bibinfo{person}{Eli Shechtman}, {and} \bibinfo{person}{Oliver Wang}.} \bibinfo{year}{2018}\natexlab{}.
\newblock \showarticletitle{The Unreasonable Effectiveness of Deep Features as a Perceptual Metric}. In \bibinfo{booktitle}{\emph{CVPR}}.
\newblock


\end{thebibliography}










\end{document}